\pdfoutput=1
\documentclass{article}
\usepackage{microtype}
\usepackage{graphicx}
\usepackage{subfigure}
\usepackage{booktabs}
\usepackage{hyperref}

\usepackage[accepted]{foobar2024}
\usepackage{amsmath}
\usepackage{amssymb}
\usepackage{mathtools}
\usepackage{amsthm}

\usepackage{multicol}
\usepackage{multirow}
\usepackage[normalem]{ulem}
\usepackage{caption}
\usepackage{subcaption}
\usepackage{listings}
\usepackage{comment}
\usepackage{adjustbox}
\usepackage{enumitem}
\newcommand{\datasetname}{\textsc{ToM-in-AMC }}
\newcommand{\datasetnamens}{\textsc{ToM-in-AMC}}
\lstdefinelanguage{prompt}{
    frame=l,
    framerule=3pt,
    framesep=8pt,
    basicstyle=\small\ttfamily,
    commentstyle=\color{cyan},
    morecomment=[l]{//},
    moredelim=[is][\color{red}\bfseries]{<<<}{>>>},
    moredelim=[is][\color{magenta}\bfseries]{[[[}{]]]},
    moredelim=[is][\color{orange}\bfseries]{===}{===},
    moredelim=[is][\color{olive}\bfseries]{|||}{|||},
}
\lstdefinelanguage{ioexample}{
    frame=shadowbox,
    rulesepcolor=\color{gray},
    framerule=0.5mm,
    rulesep=2mm,
    basicstyle=\small\normalfont,
    commentstyle=\color{cyan},
    morecomment=[l]{//},
    moredelim=[is][\color{red}\bfseries]{<<<}{>>>},
    moredelim=[is][\color{magenta}\bfseries]{[[[}{]]]},
    moredelim=[is][\color{orange}\bfseries]{===}{===},
    moredelim=[is][\color{olive}\bfseries]{|||}{|||},
    moredelim=[is][\bf]{:::}{:::},
    moredelim=[is][\it]{---}{---},
    moredelim=[is][\tt]{+++}{+++},
}
\usepackage[capitalize,noabbrev]{cleveref}
\theoremstyle{plain}

\theoremstyle{definition}

\theoremstyle{remark}

\usepackage[textsize=tiny]{todonotes}
\icmltitlerunning{Few-Shot Character Understanding in Movies as an Assessment to Meta-Learning of Theory-of-Mind}

\begin{document}

\twocolumn[
\icmltitle{Few-Shot Character Understanding in Movies\\
           as an Assessment to Meta-Learning of Theory-of-Mind}
\icmlsetsymbol{equal}{*}

\begin{icmlauthorlist}
\icmlauthor{Mo Yu}{equal,wechat}
\icmlauthor{Qiujing Wang}{equal,xjtu}
\icmlauthor{Shunchi Zhang}{equal,xjtu}
\icmlauthor{Yisi Sang}{syracuse}
\icmlauthor{Kangsheng Pu}{syracuse}
\\
\icmlauthor{Zekai Wei}{syracuse}
\icmlauthor{Han Wang}{wechat}
\icmlauthor{Liyan Xu}{wechat}
\icmlauthor{Jing Li}{njit}
\icmlauthor{Yue Yu}{lehigh}
\icmlauthor{Jie Zhou}{wechat}
\end{icmlauthorlist}

\icmlaffiliation{wechat}{Pattern Recognition Center, WeChat AI}
\icmlaffiliation{syracuse}{Syracuse University}
\icmlaffiliation{xjtu}{Xi'an Jiaotong University}
\icmlaffiliation{njit}{New Jersey Institute of Technology}
\icmlaffiliation{lehigh}{Lehigh University}

\icmlcorrespondingauthor{Mo Yu}{moyumyu@global.tencent.com}
\icmlcorrespondingauthor{Qiujing Wang}{qiujing.wang@stu.xjtu.edu.cn}
\icmlcorrespondingauthor{Shunchi Zhang}{shunchi.zhang@stu.xjtu.edu.cn}

\icmlkeywords{reading comprehension, story understanding, character understanding, theory-of-mind}

\vskip 0.3in
]

\printAffiliationsAndNotice{\icmlEqualContribution}

\begin{abstract}
When reading a story, humans can quickly understand new fictional characters with a few observations, mainly by drawing analogies to fictional and real people they already know. 
This reflects the few-shot and meta-learning essence of humans' inference of characters' mental states, \emph{i.e.}, theory-of-mind (ToM), which is largely ignored in existing research.
We fill this gap with a novel NLP dataset, \datasetnamens,
the first assessment of machines' meta-learning of ToM in a realistic narrative understanding scenario.
Our dataset consists of $\sim$1,000 parsed movie scripts, each corresponding to a few-shot character understanding task that requires models to mimic humans' ability of fast digesting characters with a few starting scenes in a new movie.

We propose a novel ToM prompting approach designed to explicitly assess the influence of multiple ToM dimensions. It surpasses existing baseline models, underscoring the significance of modeling multiple ToM dimensions for our task.
Our extensive human study verifies that humans are capable of solving our problem by inferring characters' mental states based on their previously seen movies.
In comparison, our systems based on either state-of-the-art large language models (GPT-4) or meta-learning algorithms lags $>$20\% behind, highlighting a notable limitation in existing approaches' ToM capabilities.
\end{abstract}
\section{Introduction}
\label{sec:intro}

Humans are social animals who engage in a high frequency of social activities every day. To achieve efficient social interactions, humans need to understand other people's mental states, such as intentions and beliefs, with small amounts of information and to predict their next moves~\cite{perner1985john,keysar2000taking}.
Such ability is known as theory-of-mind (ToM)~\cite{premack1978does}.

In AI research, there is also a growing interest in giving machines such theory-of-mind~\cite{nematzadeh2018evaluating,yuan2020emergence,zhu2021few}, mostly in synthetic settings.
The accomplishments of these efforts have encouraged a shift in the study of ToM from synthetic environments to real-life scenarios for potential future applications. However, the creating such evaluation benchmarks is challenging since it would be impossible to imitate human actions that take into account a variety of real-world elements.
The NLP community hence resorts to fictional characters in stories as a delegate. Characters play a central role in stories --- while reading stories, humans build mental models for characters to understand their goals, emotions, personalities, future behaviors, etc.~\cite{gernsbacher1998automatically}. 
Therefore, understanding characters in stories serves naturally as a proxy for assessing the machine's ToM ability.
Benchmarks with different task formats have been established, including \emph{personality classification}~\cite{flekova2015personality}, \emph{personalized dialogue generation}~\cite{li2020aloha}, and \emph{anonymous speaker guessing}~\cite{sang2022tvshowguess}.

All existing assessments model a character with a large amount of behavioral data and dialogues.
In contrast, humans can usually understand new people with ``\textbf{few-shot}'' observations. Instead of making judgments after accumulating observations over an extended period, when we meet strangers or read new fictional characters, we make primitive judgments based on the limited information currently available and dynamically change our impressions over time as we take in new information. 

\begin{figure*}
\centering
\includegraphics[trim=0cm 0.5cm 0cm 0cm,clip,width=0.96\textwidth]{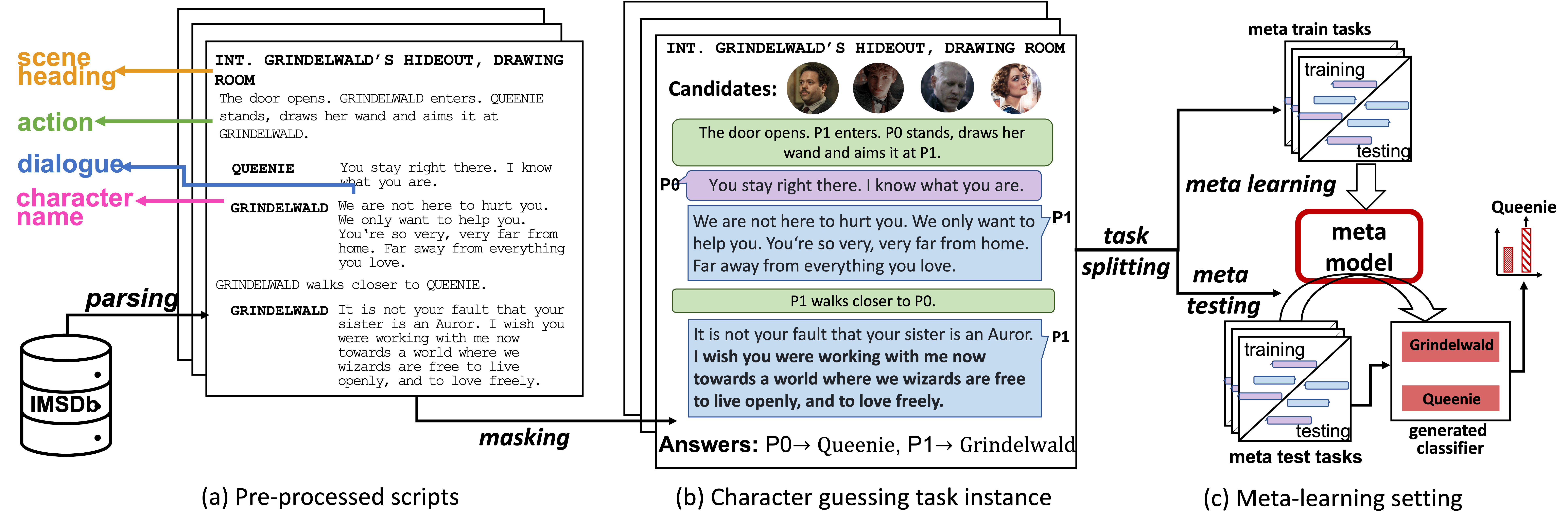}
\caption{Overview of our \datasetname task and the proposed meta-learning formulation.}
\vspace{-\baselineskip}
\label{fig:process}
\end{figure*}

Humans have this ability because of their prior experience of meeting different people and reading stories about various characters throughout their lifetime~\cite{rowe2008archetype,jahan2021inducing}, as well as employing basic cognitive functions such as association~\cite{ma2011spontaneous}.
For example, \emph{Grindelwald} in the \emph{Fantastic Beast} movies (Figure~\ref{fig:process}) is not a simple \emph{villain} but has a much more complex persona. Still, many people can easily understand his charismatic and idealistic persona as a revolutionary leader, if they have seen \emph{Magneto} from the \emph{X-men}, both of whom are revolutionary for their own kinds but are ruthlessness to others. Thus, the audience can understand \emph{Grindelwald} has the exact persona to speak the last utterance in Figure~\ref{fig:process}b.
This reflects the \textbf{meta-learning\footnote{\scriptsize{Throughout this paper we refer to its machine learning definition, focusing on generalizing to new tasks with small data~\cite{vinyals2016matching,finn2017model}.}} capability} of humans' ToM, which is largely ignored in previous works.

We aim to bridge this gap and evaluate the capacity of machines to meta-learn ToM similar to humans. Such an assessment raises two fundamental requirements for tasks.
(1) We need natural few-shot tasks that \emph{humans can effectively tackle with knowledge from related tasks}. 
To this end, we construct our dataset based on movie script understanding. When humans watch a new movie, they can rapidly comprehend the roles of unfamiliar characters based on a limited number of initial scenes from their knowledge of previously viewed movies. Here each movie naturally corresponds to a few-shot learning task, and the process of swiftly understanding new characters represents a meta-learning setting.
(2) Each task should effectively \emph{assess the ToM capabilities in a comprehensive way}.
Among the existing task formats of character understanding, we adopt the task from \cite{sang2022tvshowguess}, which requires guessing the identities of speakers in a scene with all their utterances anonymized, as shown in Figure~\ref{fig:process} (middle).
Human study illustrated that the task requires understanding multiple types of character personas that are well aligned with humans' ToM during reading.

Based on the two ideas, we created the first assessment for \textbf{meta-learning of ToM}. For each movie, the script is pre-processed (Figure~\ref{fig:process}a) to a sequence of scenes to form the character guessing instances (Figure~\ref{fig:process}b).
A small number of starting scenes sufficient for humans to grasp characters are used for training, making each movie a few-shot task.
We split the movies into meta-training 
and meta-testing tasks. A meta-model can learn from meta-training tasks, then make few-shot predictions on the meta-testing ones (Figure~\ref{fig:process}c).

\smallskip
\noindent\textbf{Transductive v.s. Inductive Settings. }
Our dataset enables the assessment of machines' ToM in two settings: a \emph{transductive} setting, where a meta-model possesses access to and can leverage the characters' previous acts as examples during prediction, and a more stringent \emph{inductive} setting, where a meta-model must generate a mental model or a mental state description of the character, relying solely on this information for prediction. 
While both settings are challenging, the inductive setting holds particular significance:

$\bullet$ \emph{Emphasizing the effects of various ToM dimensions and improving explanability}: While our task encompasses a broad range of ToM dimensions through the use of real-world scenarios, it lacks specific evaluations tailored to each dimension. The inductive approach offers the opportunity to model each dimension using the generated mental descriptions, allowing for explicit performance measurement and in-depth study of their respective impacts.

$\bullet$ \emph{Mitigating Shortcuts}: The existing ToM assessments primarily focus on the end-task performance. This carries the risk that machines may achieve correct predictions through shortcuts, e.g., data leaks during pre-training or spurious correlations~\cite{shapira2023clever} of non-ToM-related cues. Requiring to generate mental descriptions can help alleviate such shortcuts and lead to a more effective evaluation.

\smallskip
\noindent\textbf{Main Observations. }
We conduct a large-scale human study on our dataset.
The human annotators are asked to perform the tasks on \emph{movies they have not seen before}.
The results show that they can solve our task with a $\sim$90\% accuracy, with the help of their knowledge acquired from their previously seen movie characters. 
In comparison, the widely used prototypical networks~\cite{snell2017prototypical} and the LEOPARD~\cite{bansal2019learning} lag $\sim$30\% behind humans.
We also investigate the usage of large language model (LLM) GPT-4. Our proposed \textbf{ToM prompting}, which explicitly models the separated mental states, including belief, intention, personality, and more, achieves the best performance, confirming the effectiveness in modeling various dimensions of ToM for our task. Nevertheless, it is $>$20\% behind humans, emphasizing the significant challenge presented by our task and the substantial advancements that AI systems still need to achieve to fully grasp ToM.

Our work makes the following contributions:

(1) We propose the problem of few-shot character understanding to assess machines' meta-learning ability of theory-of-mind, which is common in humans' daily life but has been overlooked by AI and NLP research; and build the first dataset to this end in a non-synthetic scenario.

(2) We benchmark existing meta-learning approaches and conduct comprehensive human study on our dataset, revealing that humans solve our problem with meta-learning-style strategies and largely outperform all the AI methods.

(3) We propose a novel prompting approach for the inductive setting. It outperforms existing baselines and confirms that our task requires multiple ToM dimensions to solve.
\section{Related Work}
\label{sec:related}

\noindent\textbf{Theory of Mind in NLP. }
Researchers in the NLP field have proposed several tasks to evaluate machines' ToM in the language understanding setting~\cite{ma2023towards}, particularly in light of the promising advancements seen in LLMs' emerging ToM capabilities~\cite{kosinski2023theory,bubeck2023sparks}.
Most of these tasks conduct assessments on the ToM dimension of belief~\cite{nematzadeh2018evaluating,cohen2021exploring,he2023hi,sileo2023mindgames,shapira2023clever}, which some also cover other dimensions like intention, desire, emotion, etc~\cite{zhang2010towards,sap2019social,yuan2020emergence,zhu2021few,tracey2022best,zhou2023i,wu2023coke}.

Different from our work, these datasets primarily rely on synthetic settings.
The drawback of synthetic settings is that the roles evaluated in the datasets \textbf{lack clear associations with specific characters}, which leads to a significant oversight from the crucial meta-learning perspective of ToM. Additionally, this limitation prevents the exploration of certain vital aspects of ToM, such the influence of characters' personalities and their past experiences on ongoing events.

\smallskip
\noindent\textbf{Fictional Character Understanding. }
There are many tasks proposed for character understanding in stories, covering the assessments of factual information of characters~\cite{chen2016character,chen2017robust}, inter-character relationships~\cite{massey2015annotating}, and personality of characters~\cite{flekova2015personality}.
Recent work~\cite{brahman2021let,sang2022tvshowguess} proposed a new
character guessing task --- a form of guessing the identity of anonymized characters in a scene.
Human studies showed that the task requires understanding multiple dimensions of characters' mental states, such as personalities, desires and intentions.
Therefore, our work chooses this form due to its simplicity, comprehensiveness and assessment strength.

\smallskip
\noindent\textbf{Meta and Few-Shot Learning in NLP. }
Most of the meta and few-shot learning datasets
have their tasks sampled from a single large dataset, leading to homogeneous settings.
FewRel~\cite{han2018fewrel} downsamples a relational classification dataset. SNIPS~\cite{coucke2018snips} and CLINC150~\cite{larson2019evaluation} downsamples intent classification datasets from a few general domains. 
To encourage meta-learning across heterogeneous tasks, people build datasets that collect tasks from diverse resources.
Crossfit~\cite{ye2021crossfit} collected and down-sampled 160 NLP tasks from Huggingface Datasets.
FewJoint~\cite{hou2020fewjoint} include slot-filling tasks from 59 domains. 
\citet{yu2018diverse} collect clients' proposed intent classification tasks. 
Compared to these prior work, our dataset has a natural few-shot learning setting from daily life that does not need artificial construction like down-sampling.
\begin{table}[t!]
\small
\centering
\caption{\small{Movie genres in our dataset.}}
\vspace{-0.1in}
\label{tab:genre}
\begin{adjustbox}{width=\linewidth}
\begin{tabular}{lrr||lrr} 
\toprule
\bf  Genre& \bf Count& \bf Example & \bf  Genre& \bf Count& \bf Example\\
\midrule
 Action & 201& \emph{Rush Hour2} & Horror& 99&\emph{Carrie} \\
 Adventure& 102& \emph{Tropic Thunder} & Musical& 12&\emph{Nine}\\
 Animation& 21&\emph{Toy Story} & Mystery& 69& \emph{Rear Window} \\
 Comedy& 233& \emph{Extract} & Romance&122 & \emph{Blue Valentine}\\
 Crime& 147&\emph{Deception} & Sci-Fi& 105& \emph{Jurassic Park} \\
 Drama& 394&\emph{Fracture} & Short& 2& \emph{Quantum Project}\\
 Family& 17&\emph{Up}& Thriller& 257 &\emph{Chasing Sleep}\\
 Fantasy& 66&\emph{Watchmen}& War& 15&\emph{Platoon}\\
 Film-noir& 4&\emph{Sunset Blvd.}&Western& 7& \emph{Roughshod}\\
\bottomrule
\end{tabular}
\end{adjustbox}
\vspace{-0.1in}
\end{table}

\begin{table*}[t!]
\small
\centering
\caption{\small{Statistics of our \datasetnamens.}}
\vspace{-\baselineskip}
    \renewcommand{\arraystretch}{1}
    \begin{tabular}{l ccc cc cc} 
        \toprule
        \multirow{2}{*}{\bf Task Set} & \multirow{2}{*}{\bf \#Movies} & \multirow{2}{*}{\bf \#Characters}  & \multirow{2}{*}{\bf \#Scenes} & \multicolumn{2}{c}{\bf Training Data} & \multicolumn{2}{c}{\bf Testing Data}\\ 
        \cmidrule(lr){5-6}
        \cmidrule(lr){7-8}
        &&&&{\bf \#Scenes}&\bf \#Instances&{\bf \#Scenes}&\bf \#Instances\\
        \midrule
        \texttt{Training} &807 &3,063&59,301&36,662&59,743&22,639&35,537\\
        \texttt{Development} &100 &401 &7,430&4,544&7,609&2,886&4,733 \\
        \texttt{Testing} &100 &373&7,293&4,538&7,158&2,755&4,266\\
        \midrule
        total&1,007&3,837&74,024& 45,744& 74,510 & 28,280 &44,536 \\
        \bottomrule
    \end{tabular}
    \vspace{-0.1in}
    \label{tab:dataset_stats}
\end{table*}

\section{Problem Definition}
\label{sec:problem}

To provide an assessment to the machine's meta-learning ability of ToM, we propose to mimic the scenario where humans can quickly understand characters in a new movie based on movies they have seen before.
For each movie, we build a character guessing task~\cite{sang2022tvshowguess} (Section~\ref{ssec:basic_task}), which has been verified as a valid ToM assessment.
We build our meta-setting on top of this task in Section~
\ref{ssec:problem}.

\subsection{Background: Meta-Learning Formulation}
In a meta-learning problem, we are given $N$ tasks $\mathcal{T} = \left \{ T_1, \cdots, T_N \right \}$, divided into training, development and test task sets $\mathcal{T}^{train}$, $\mathcal{T}^{dev}$, $\mathcal{T}^{test}$. 
Each task $T_i$ consists of a training data set $\mathcal{D}^{train}_i$ and a test data set $\mathcal{D}^{test}_i$. 

A typical meta-learning model consists of two stages. (1) A \textbf{meta-training} stage 
learns a meta-model on the few-shot tasks in $\mathcal{T}^{train}$.
In each iteration, a task $T_i$ is sampled from $\mathcal{T}^{train}$. The meta-model trains on samples from $\mathcal{D}^{train}_i$ and is tested on samples from $\mathcal{D}^{test}_i$.
The testing loss is used to optimize the meta-model's parameters. 
Its hyperparameters are determined with the meta-dev set $\mathcal{T}^{dev}$.
(2) A \textbf{meta-testing} stage evaluates the learned meta-model on the unseen tasks from $\mathcal{T}^{test}$, which typically outputs a classifier by adapting on its small number of training samples.
The ultimate goal of a meta-learning is to efficiently transfer the knowledge about learning on the training tasks to new tasks.

\subsection{Background: Character Guessing Task}
\label{ssec:basic_task}
Each of our tasks has the character guessing format~\cite{sang2022tvshowguess}.
The task adopts a multi-choice setting.
The input is a scene with the main characters (at most 5 for each movie) masked with their corresponding IDs. The IDs are randomly assigned to characters in different scenes.
The goal is to map each ID to its identity.
Formally, we denote the $t$-th anonymous scene in a movie as $\mathcal{S}^{(t)}=\{s^{(t)}_1, s^{(t)}_2, ..., s^{(t)}_n\}$.
$s^{(t)}_i$ is an utterance or background description, which depicts the verbal or behavioral actions of anonymous characters with ID $\text{P}_x$, $x \le 5$. A scene is associated to a candidate character set $\mathcal{C}= {c_1,...,c_k}$, $k \le 5$.
The goal is thus to predict each $\text{P}_x$'s actual identity $c^{(t)}_j$ as:
\begin{equation}
\small
\setlength{\abovedisplayskip}{4pt}
\setlength{\belowdisplayskip}{2pt}
\begin{aligned}
    P(\text{P}_x = c^{(t)}_j\vert {\mathcal{S}}^{(t)}).
\end{aligned}
\end{equation}

\subsection{Meta-Learning of Character Guessing}
\label{ssec:problem}
Different from \cite{sang2022tvshowguess} where each character has a large amount of training data, our work has a natural few-shot setting.
We have a set of movie scripts $\mathcal{M} = \{M_1, \cdots, M_N\}$. Each movie $M_i$ corresponds to a task $T_i$ in the meta-learning formulation.
The main characters in each movie $M_i$, denoted as $\mathcal{C}_i$, are treated as class labels.
Each instance of $T_i$ is a tuple $(\textrm{P}_x=c_k, S)$, where $S$ is an anonymous scene from $M_i$. $\textrm{P}_x$ is one of the masked characters in $S$, which has the actual identity of $c_k$.

For each $M_i$, we split a few starting scenes into the training set, which are sufficient for human to grasp characters.
The problem asks a meta-model to learn from training movies, so as to perform well on unseen movies with few-shot examples.
In this way, it assesses how to infer a new character's mental states rapidly by drawing analog from seen characters, \emph{i.e.}, the meta-learning ability of ToM.
\section{Our \datasetname Benchmark}
\label{sec:dataset}
We constructed \datasetnamens, the first dataset on \underline{ToM} {meta-learn\underline{in}g} \underline{A}ssessment with \underline{M}ovie \underline{C}haracters as a testbed. We collect movie scripts from IMSDB (\url{imsdb.com}), divide the script into scenes, and recognize and anonymize the main characters in each scene. Finally, we build a task on each movie to simulate few-shot scenarios. In total, we collected 1,007 movies. Table~\ref{tab:genre}
shows that \emph{Drama}, \emph{Thrill}, and \emph{Comedy} are the 3 most popular genres.

\smallskip
\noindent\textbf{Script Parsing and Scene Splitting. }
Movie scripts are highly structured documents that have basic formatting elements~\cite{riley2009hollywood}, as shown in Figure~\ref{fig:process}(a), 
including (1) \textbf{scene headings} that indicate the start of a scene with place information; (2) \textbf{actions} that describe the characters' behaviors and the setting; and (3) \textbf{dialogues} of the characters. 

\begin{figure*}[t!]
\setlength{\belowcaptionskip}{-\baselineskip}
\setlength{\abovecaptionskip}{.4\baselineskip}
\centering
\includegraphics[trim=0cm 0.05cm 0cm 0.1cm,clip,width=.88\textwidth]{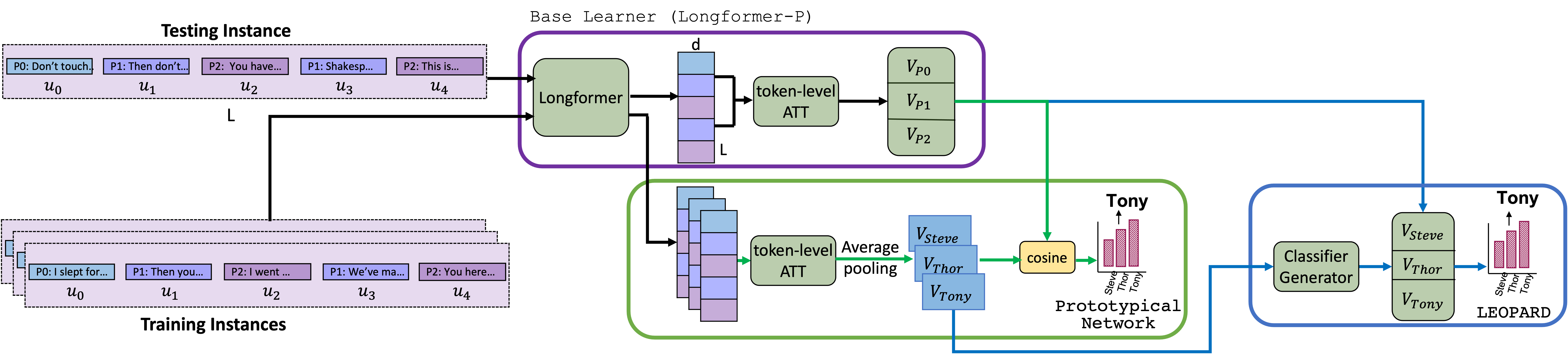}
\caption{\small{Our two proposed meta-learning approaches for the character prediction task. (top) the base learner (Longformer-P); (middle) the prototypical network approach; (right) the LEOPARD approach.}}
\label{fig:model}
\end{figure*}

We process the scripts with a state-of-the-art parser from~\cite{sang2022mbti} to identify headings, actions, and dialogues; then split the identified sequence of chunks into scenes according to the recognized scene headings.
Since the scene headings always first illustrate if the scene is indoors (INT.) or outdoors (EXT.), they can be accurately identified with rules.
The texts, including actions and dialogues, between two headings are considered as one scene.

\smallskip
\noindent\textbf{Evaluation Task Construction. }
We choose the top-5 characters with the most dialogue utterances as candidates for each movie, so that each has sufficient evidence for our character identification task. 
We use the first 3/5 of the movie script for training and the rest for testing. 
According to \cite{NC}, in movie scripts, the main characters are usually introduced in the first 10 pages with their personalities and appearances, to provide a mental picture for the readers.
Therefore, our training split is able to cover sufficient information for humans to understand characters.
In our problem, we denote each character in a scene as an instance.
As shown in Table \ref{tab:dataset_stats}, 
every character has less than 20 training instances on average, naturally leading to a few-shot problem setting.

\smallskip
\noindent\textbf{Name Perturbation. }
The LLMs have a vast amount of pre-training data, including some of the testing movies.
This leads to data leak thus powerful LLMs like GPT-4 may resolve our task with its memorization instead of ToM reasoning.
To mitigate this issue, we introduced perturbations in our testing tasks by replacing main character names with random English names while preserving their genders (details in Appendix~\ref{app:perturbation_setting}). This prevents the model from relying on memorized movie information. Non-LLM results are unaffected by these perturbations since they treat names as class labels, and human annotators had not seen the movies.
\section{Baseline Methods}
\label{sec:method}

We introduce the baselines adapted to our problem in the
transductive and inductive settings, according to whether the method explicitly produces a mental model or a mental state of a character. 

\subsection{Transductive Learning Approaches}

\paragraph{Prototypical Network~\cite{snell2017prototypical}.} The method learns a metric network $\Lambda$ for prediction. 
In our work, $\Lambda$ is a base learner, Longformer-P (model architecture detailed in Appendix~\ref{app:base_learner}), that produces embedding vectors of characters contextualized by the input scenes.
For any input pair $(x,x')$, $\Lambda(x)^T\Lambda(x')$ outputs a similarity score.
During prediction, there is not a specific model for a character. The prediction is achieved based on the similarity between the input and the characters' historical scenes (Figure~\ref{fig:model} (middle)). Therefore, it is a standard transductive learning method.
The detailed implementation can be found in Appendix~\ref{app:proto_net}.

\paragraph{In-Context Learning with LLMs.}
Large language models have demonstrated their in-context learning (ICL) capability~\cite{brown2020language}. 
This approach naturally aligns with our few-shot learning task, where previous scenes from the training sets can be utilized as few-shot demonstrations to aid in making predictions for testing scenes.
In our study, we use the ChatGPT and GPT-4 as the LLMs. 
Considering the maximum input lengths permitted by the model services, we include 10 or 20 demonstrations (referred to as \emph{10-shot} or \emph{20-shot}) along with a testing case for predictions. The detailed prompt construction can be found in Appendix~\ref{app:llm_prompt}.

\begin{figure*}[t!]
    \centering
    \includegraphics[width=0.94\textwidth]{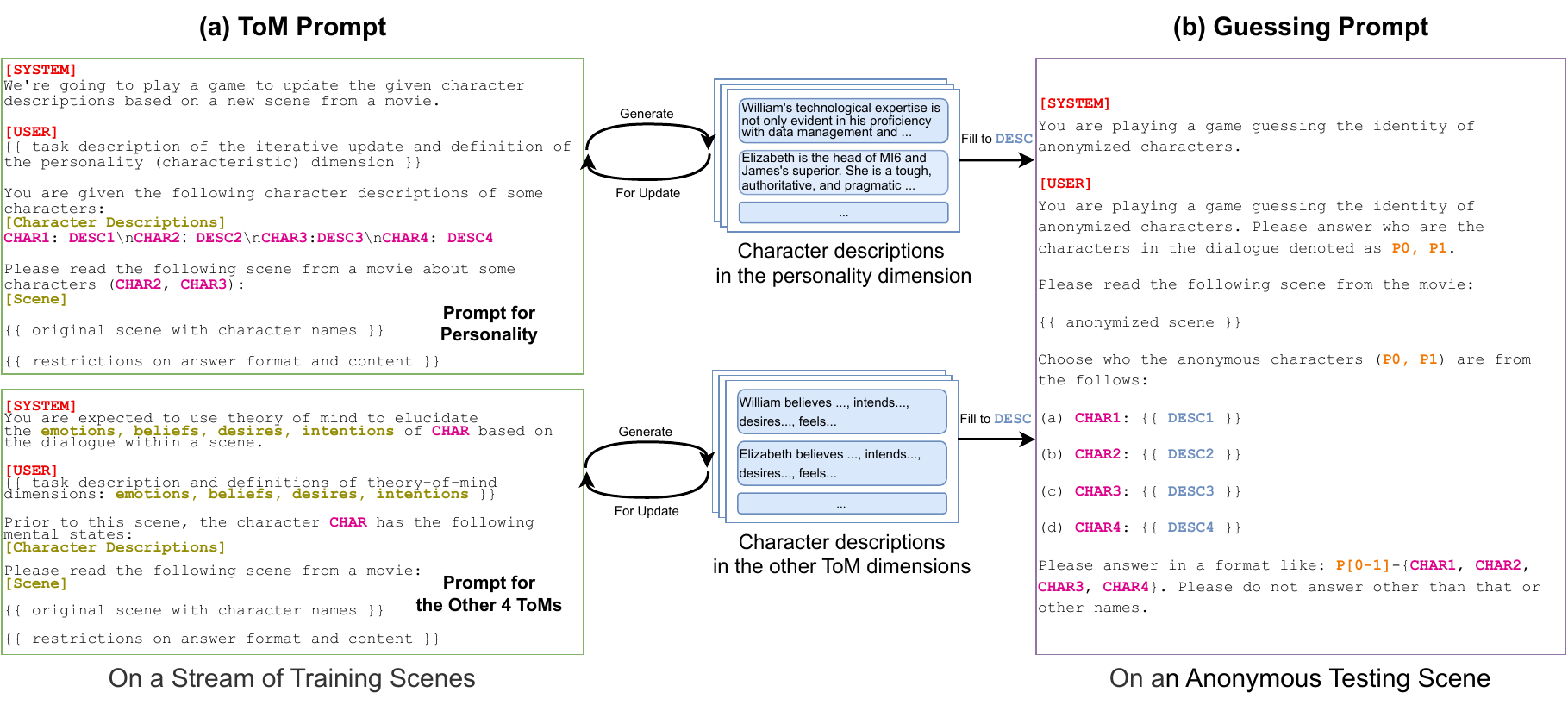}
    \vspace{-0.1in}
    \caption{Our proposed ToMPro approach. The method first (a) generates character mental descriptions along multiple ToM dimensions based on input scenes; then (b) predicts the identities of a new testing scene with the generated descriptions.}
    \vspace{-0.15in}
    \label{fig:llm-methods}
\end{figure*}

\subsection{Inductive Learning Approaches}

\smallskip
\noindent\textbf{Multi-Task Learning.}
A most straightforward inductive baseline is to apply standard multi-task learning on all the training and evaluation tasks to learn a classifier for each character.
All the tasks share the same Longformer encoder, \emph{i.e.}, the base learner in Appendix~\ref{app:base_learner}.
On top of the encoder, each task $i$ has its own prediction layer $f_i$, which is a linear classification head that makes prediction as $P(\text{P}_x=c|S) =  f_{i}(\mathbf{e}_{\text{P}_x|S})$, where $f_{i}: \mathbb{R}^{d} \xrightarrow{} \mathbb{R}^{C}$.
We do early stopping according to the averaged performance across \emph{the testing data of all development tasks} for model selection.

\smallskip
\noindent\textbf{LEOPARD~\cite{bansal2019learning}.}
The LEOPARD algorithm is originally introduced to handle the challenge of varying numbers of classes across tasks in few-shot learning.
Compared to the standard MAML~\cite{finn2017model} algorithm, it consists of an additional parameter generator, which learns to generate the initial parameters of the prediction layer for a new task. 
Therefore, the algorithm is able to output a model of each character for prediction and becomes an inductive approach.
The details about how we adapt the LEOPARD to our problem can be found in Appendix~\ref{app:leopard}.
\section{The ToM Prompting Method}
\label{sec:tompro}
We introduce \textbf{ToMPro}, an LLM-based inductive learning method with two prompting stages.
In Stage-1 (Figure~\ref{fig:llm-methods}a), it iteratively analyzes a stream of scenes to update character mental states across various ToM dimensions: personality, emotions, beliefs, desires, and intentions. The prompt uses the current scene and characters' prior mental state descriptions for reference.
In Stage-2 (Figure~\ref{fig:llm-methods}b), LLMs identify anonymized characters in test scenes based on descriptions obtained in the first stage.
The two stages correspond to the two major functions of ToM, \emph{understanding} others' mental states and \emph{reasoning} about their future behaviors with the knowledge about their mental states.

We tune the format and phrasing of the prompts on the TV show transcripts from~\cite{sang2022tvshowguess}, to prevent overfitting to our movie scripts.
The final prompt for Stage-1 incorporates the following adjustments.
First, we use a separate prompt to simultaneously generate the personality dimension for all characters in a scene. This decision is based on the observation that generating personality descriptions separately for each character is likely to yield homogeneous outcomes due to the limited variability within this dimension.
Second, we observed that the personality dimension heavily relies on previous step results, because the character's personality tends to remain stable but needs to be gradually revealed as the story progresses.
In contrast, the other ToM dimensions generally represent short-term states and predominantly depend on the current scenes.
Appendix~\ref{app:tompro} gives the detailed prompts for the two stages.
\section{Experiments}
\label{sec:exp}

\subsection{Baselines and Implementation Details}
\label{ssec:exp_details}

We evaluate the instance-level accuracy. An instance is a masked speaker in a scene.
We implement the non-LLM baselines based on HuggingFace~\cite{wolf-etal-2020-transformers}, with the \texttt{allenai/longformer-base-4096} for initialization.
We optimize the models with Adam.
The LEOPARD starts with the encoder of the trained prototypical network. We train our model on a single V100 GPU. It takes around 2 hours to train one epoch. We train the MTL baseline and Prototypical Network baseline for 20 epochs and train the LEOPARD with 10 epochs.
To generate the LLM results, we use the \texttt{gpt-3.5-turbo-0613} and \texttt{gpt-4-1106}.

\smallskip
\noindent\textbf{Hyperparameters.}
We set the maximum length to 2000, which can handle most of the scenes. We set window size to 256 and batch size to 8. 
We set learning rate to 2e-5 for the MTL and prototypical network and update every 8 batches. For LEOPARD, the learning rate is 1e-5; and the parameters are updated after each inner-loop. For each model, we ran twice and found the average development accuracy varies by less than 1\%. Hence, we report our results with a single run.
For GPT-based methods, we set the temperature to 0.1 and take the average of 3 runs for evaluation.
The temperature is set to 1.0 in Stage-1 of ToMPro for diverse generations.

\subsection{Main Results I: Human Performance}
\label{sec:humanEval}
To understand the properties of \datasetnamens, we conduct a human study on the development set. Specifically, we explore (1) \emph{the human performance} on our task and (2) \emph{the dependency on the historical events} to complete our task. 

We sampled 11 movies from the genres with script counts greater than 100 in the development set. 335 scenes from the 11 movies are distributed to two raters who have \emph{not} watched the movies. 
The raters perform two tasks on each scene: (1) guessing the character identities and (2) identifying whether the guessing task needs only the current scene or additional scenes from the movie. 
We evaluate the instance-level accuracy of the raters, where an instance refers to a masked speaker in a scene. For each instance, raters have five options who are the main characters. Figure~\ref{fig: human_study_interface} in Appendix~\ref{app:interface} shows our human study interfaces. In total, the raters annotated 569 instances in these scenes. 

\smallskip
\noindent\textbf{Results. } 
Humans can solve our tasks quite well, with an average human performance of 88.0\%. As will be shown in the experimental section, it largely outperforms the model performance by {$>$20\%}, showing a significant gap for machines to improve.
Many human errors come from hard or unsolvable cases (such examples are shown in Appendix~\ref{app:human_error}).

Importantly, our study shows that there is no significant shortcut for guessing the characters without persona understanding. Raters often need to read the whole movie script before the scene to understand the characters' personae. There are 311 scenes that require historical scenes to resolve, corresponding to 92.84\% of the examples.

\smallskip
\noindent\textbf{Remark on Humans' Solutions. } Our study also revealed that to solve our task, humans frequently leverage their knowledge from seen movies, which corresponds to a ``meta-learning'' style solution.
Specifically, the raters reported the following strategies they used to understand a new character:
(1) They first classify the characters to rough \textbf{archetypes} they learned from previous experience, \emph{e.g.}, \emph{Hero} and \emph{Villain}.
(2) When archetypes are insufficient,
they \textbf{associate} the new characters with the ones in movies they have seen before, to make a fine-grained understanding.
Appendix~\ref{app:human_solution} provides examples of this association process.

\begin{table}[t!]
\setlength{\belowcaptionskip}{-\baselineskip}
\setlength{\abovecaptionskip}{0.5\baselineskip}
    \small
    \centering
    \vspace{0.2in}
    \caption{\small{Overall
    performance (\%) on our \datasetname task.
    (*) Evaluation was conducted on a subset of the dataset (see Appendix Table~\ref{tab:dev_sampled_movies} and \ref{tab:test_sampled_movies}). $\dagger$ the dataset released by \cite{sang2022tvshowguess}.}}
    \renewcommand{\arraystretch}{1}
    \begin{tabular}{lcccc} 
        \toprule
        \bf System & \bf Dev Acc  & \bf Test Acc  \\ 
        \midrule
        Random & 22.1 &  25.0  \\
        Majority & 34.9  & 36.0  \\
        Human$^{*}$ & 88.0 & -- \\
        \midrule
        \midrule
        \multicolumn{3}{c}{\emph{Transductive Setting}}\\
        Proto. Net & 55.4 & 53.2 \\
        \quad - Trained on TVSG$^\dagger$&45.9 & 46.4 \\
        GPT-4 ICL (20-shot)$^{*}$ & \bf 67.8 & -- \\
        \quad - 10-shot & 63.8 & --\\
        \quad - replaced with GPT-3.5 (10-shot)$^{*}$  &      54.9   & --   \\
        \midrule
        \midrule
        \multicolumn{3}{c}{\emph{Inductive Setting}}\\
        MTL of Classifiers                 & 42.8     & 38.1 \\
        LEOPARD                            & 59.4     & 58.6 \\
        \midrule
        GPT-4 ToMPro$^{*}$      &     \bf 68.2     & 66.9   \\
        \quad - w/o update after training scenes$^{*}$  &      61.8    & --   \\
        \quad - replaced with GPT-3.5$^{*}$  &      60.3    & --   \\
        \bottomrule
    \end{tabular}
    \vspace*{-0.2in}
    \label{tab:overall_performance}
\end{table}

\subsection{Main Results II: Machine Performance}

\paragraph{Transductive Setting.}
The middle part of Table~\ref{tab:overall_performance} compares different models in the transductive setting. The prototypical network trained on our \datasetname achieves 55.4\%, significantly better than the random and majority baselines.
To confirm the value of our training data, we directly use the TVSG model from \cite{sang2022tvshowguess} as the prototypical network. Its inferior performance confirms the diversity among fictional characters, showing the limitations of prior work that relies on large data per character and justifying the importance of studying the meta-learning setting for character understanding.
GPT-4 ICL approach achieves respectable performance on our task, which performs 12\% higher than the best prototypical network result. 

\paragraph{Inductive Setting.}
The bottom of Table~\ref{tab:overall_performance} compares different approaches in the inductive setting, where each approach explicitly builds a model or a mental state description for a character.
For the non-LLM methods, the multi-task learning (MTL) baseline suffers from limited training data and performs poorly; but the LEOPARD outperforms the prototypical network and becomes the best non-LLM baseline.

\begin{figure}
    \centering
    \vspace{-0.1in}
    \includegraphics[width=0.85\linewidth]{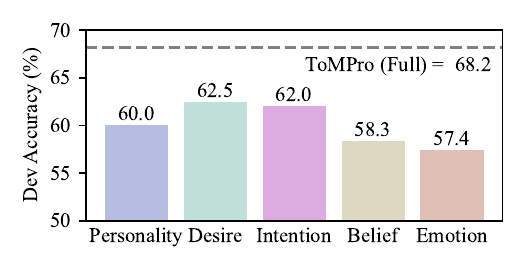}
    \vspace{-0.2in}
    \caption{\small{Ablation of ToMPro on the 5 ToM dimensions.}}
    \vspace{-0.2in}
    \label{fig:5dim}
\end{figure}

Our ToMPro approach significantly outperforms all other inductive baselines. The ablation study shows that (1) all the ToM dimensions contribute to the improvement (Figure~\ref{fig:5dim}), affirming that our task necessitates a comprehensive understanding of ToM. Among the dimensions, desire and intention are most crucial for our task, while emotion is the least crucial among the five; (2) our iterative approach enables to utilize the immediate history of a testing scene, extending beyond the usage of training scenes alone. This feature is crucial for enhancing the quality of short-term mental states (61.8\% to 68.2\%).

Despite ToMPro's success in generating characters' mental representations, the performance still significantly lags behind human by $\sim$20\%.
It shows that the LLMs are still far from reaching human-level ToM, and suggests the great potential for future improvement.
Through a qualitative analysis of ToMPro's generated descriptions (see Appendix~\ref{app:tom_examples} for examples and detailed discussions), we make the following observations:
(1) In the descriptions, ToMPro tends to include the key evidence events it uses to infer the mental states, which offers a substitute representation to original scripts and helps surpass all transductive methods;
(2) ToMPro struggles to generate high-quality desires, due to GPT-4's limited global picture of characters;
(3) Error propagation occurs when using historical states as inputs, calling for improvements to utilize historical information for robust generation.
(4) ToMPro often includes trivial facts, as GPT-4 struggles to distinguish significant information, leading to misleading character depictions.
(5) Even a generated description is accurate, ToMPro may make mistakes during the guessing stage when contextual coherence between immediate mental states and the current scene is lacking. This reflects the deficiency in ToM reasoning of LLMs and calls for enhancement of LLMs to develop global understanding and abstraction of historical mental states.

\subsection{Analysis}
\label{ssec:analysis}

\begin{figure*}[ht]
\begin{minipage}[c]{0.64\textwidth}
\begin{table}[H]
    \caption{\small{Performance by difficulty levels measured the number of speakers in a scene.}}
    \vspace{-0.7\baselineskip}
    \setlength{\tabcolsep}{2.8pt}
    \begin{tabular}{cc cc cc c}
\toprule
     \multirow{2}{*}{\textbf{Difficulty}} &
     \multirow{2}{*}{\textbf{\#Speakers}} &
     \multicolumn{2}{c}{\bf Transductive} &
     \multicolumn{2}{c}{\bf Inductive} &
     \multirow{2}{*}{\bf Human$^*$} \\
     \cmidrule(lr){3-4} \cmidrule(lr){5-6}
     &
     & \textbf{ProtoNet}
     & \textbf{ICL$^*$}
     & \textbf{LEOPARD}
     & \textbf{ToMPro$^*$}
     \\
\midrule
Easy           & $<$ 3         & 56.0 & 72.0 & 63.1 &  70.3  & 89.5 \\
Hard           & $\geqslant$ 3 & 47.7 & 57.7 & 48.6 & 62.9 & 84.2 \\
\midrule
$\Delta$       &               &  8.3 & 14.3 & 14.4 &   7.4    &  5.3 \\
\bottomrule
\end{tabular}
    \label{tab:breakdown_difficulty}
\end{table}
\end{minipage}
\hfill
\begin{minipage}[c]{0.35\textwidth}
    \includegraphics[width=0.95\linewidth]{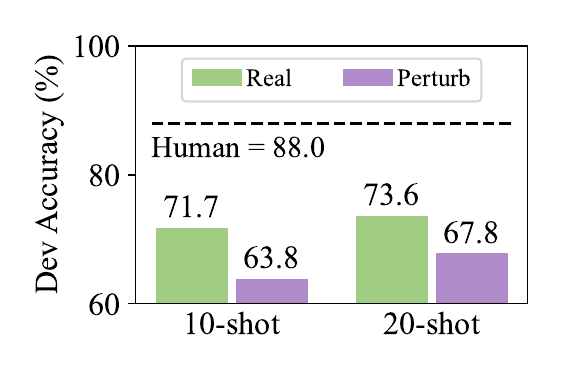}
    \vspace{-0.2in}
    \caption{\small{Effects of perturbation on GPT-4 ICL.}}
    \label{fig:perturbation_results}
\end{minipage}
\vspace{-4mm}
\end{figure*}

\begin{figure}[t!]
    \centering
    \vspace{-0.1in}
    \includegraphics[width=0.42\textwidth]{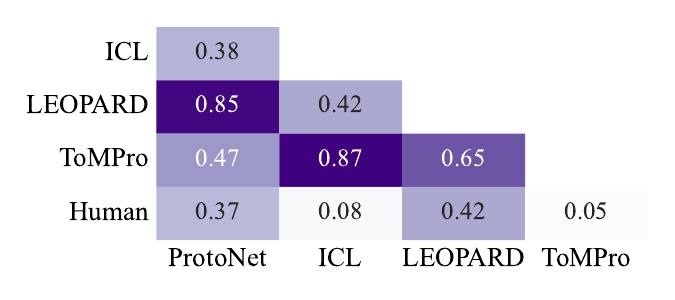}
    \vspace{-0.2in}
    \caption{\small{Correlation between models across genres.}}
    \vspace{-0.2in}
    \label{fig:corr_mat}
\end{figure}

\smallskip
\noindent\textbf{Performance on Different Numbers of Choices.}
We investigate the dependency between accuracy and the number of characters contained in the scene to guess. Table~\ref{tab:breakdown_difficulty} gives the performance decomposition according to if a scene consists of $\geqslant 3$ characters (detailed performance breakdown to the number of choices in Table \ref{tab:number_choices_results}, Appendix~\ref{app:addditional_exp}).
As expected, the involvement of multiple speakers has a noticeable impact on all the evaluated approaches, primarily due to the limited evidence per speaker and the increasing complexity of the conversational logic.
Furthermore, scenes with fewer speakers can sometimes be solved with shortcuts, \emph{e.g.}, exploiting correlations between locations, genders, and characters.
In contrast, humans employ their ToM capabilities to tackle our tasks, consistently delivering comparable performance levels across scenes with varying numbers of options.

Our ToMPro leads to a smaller gap between the two sets while maintaining top performance, indicating that our approach relies more on ToM reasoning rather than shortcuts.

\smallskip
\noindent\textbf{Performance on Movie Genre.}
We analyze whether certain genres raise more challenges in our task.
We find that different approach categories show clear discrepancies in their performance among different genres.
Figure~\ref{fig:corr_mat} gives the Spearman correlation coefficient matrix between models across different genres.
It shows that the non-LLM approaches, LLM approaches and humans use very different strategies to reason the character identities.
Detailed performance breakdowns of movie genres are in Table \ref{tab:breakdown_genre}.

\smallskip
\noindent\textbf{Effects of GPT-4's Memorization. }
To gain a deeper understanding of GPT-4's memorization issue and the necessity of our perturbation setting, we conducted an analysis using the original non-perturbed data to compare the results.
First, we devised a zero-shot experiment in which we asked GPT-4 to identify characters solely based on their names as options, without any historical context or character descriptions. This experiment resulted in an accuracy of 69.2\%, which indicates that GPT-4 has indeed been extensively exposed to the content of our movies during its training.
Second, we compared the performance of our GPT-4 ICL approach in both the perturbed and non-perturbed settings. Figure~\ref{fig:perturbation_results} shows a significant gap in their results. These results suggest that our perturbation setting effectively enhances the evaluation of ToM abilities by mitigating the impact of memorization.

\smallskip
\noindent\textbf{Mental State Generation w/ and w/o GPT-4's Memorization. }
Continuing from the previous analysis, we delve deeper into the impact of GPT-4's memorization in our inductive setting. To facilitate a direct comparison to our ToMPro, this corresponds to generating the mental states (Stage-1) on the non-perturbed scenes and guessing the identities (Stage-2) in the perturbed scenario.

First, we substitute the mental states produced by ToMPro with \emph{GPT-4's recollection of the characters' persona}. The prompt we used can be found in Appendix~\ref{app:mem_pro}. It gives a score of 67.4. 
This method is akin to cheating because the description often includes spoilers to our testing scenes. In light of this, our ToMPro still gives better result (68.2\%), highlighting the crucial role of generating a robust mental representation of characters in our task.

Second, employing the \emph{mental states generated by ToMPro on the non-perturbed scenes} results in a small improvement from 68.2\% to 70.7\%.
It shows that the mental state generation stage is less affected by shortcuts and memorization, highlighting the significance of our inductive setting. 
The fact that LLMs are still far from humans' ToM capabilities even with data leaks from real character names shows the great potential for future work.

\smallskip
\noindent\textbf{Does GPT-4 Have Correct Understanding of the ToM Dimensions? }
To ensure ToMPro accurately understands the definitions of various ToM dimensions and produces mental descriptions for the required dimensions, we perform human verification on 280 generated cases. 
It reveals that humans can recognize the dimension from GPT-4's mental descriptions 94\% of the time. However, GPT-4 often struggles to generate long-term desire descriptions due to its limited big-picture understanding, leading to correlated desires and intentions (still distinguishable through expressions).
\section{Conclusion}
\label{sec:conclu}
Inspired by the fact that humans can quickly infer the mental states of fictional characters when seeing a new story, we present the problem of studying machines' ability in meta-learning of ToM and a benchmark for this assessment.
Our experiments and human study justify the value of our benchmark, as
(1) humans greatly outperform all the meta-learning approaches including the GPT-4 based ones on our dataset with a $\sim$20\% margin;
(2) human solve our task largely with the knowledge about characters obtained from the stories they have read before;
(3) our proposed ToMPro method demonstrates that our task benefits from the understanding of multiple ToM dimensions.
Our work suggests the great value and potential for future study to fuel machines with ToM via meta-learning.
\section*{Impact Statement}

\paragraph{Limitations}
Our proposed inductive setting and the perturbation setting have successfully addressed some inherent challenges in studies involving pre-trained language models. Specifically, there is likely to be testing data leaks during LLMs' pre-training stage, especially when the testing examples are from a publicly available and widely known domain.
Also the powerful methods, like in-context learning of GPT-4, may make use of spurious correlations as shortcuts to solve the problems. In our case, the location where a scene happens and the characters who are more likely to appear in the locations makes a such shortcut.

However, it is important to acknowledge that our methodology still has certain limitations --- due to the imbalance in the number of characters across genders, the female characters can always be correctly identified without the need for historical information. 
In the future, this imbalance issue can be addressed by manually labeling a small subset of movies with balanced character representation across genders. This would allow for name perturbation while ensuring gender parity.

There is a limitation to our ToMPro approach that should be noted. While it has shown remarkable performance and holds significant promise for research, it is dependent on powerful LLMs like GPT-4, which can be resource-intensive. In contrast, lighter-weight LLMs such as GPT-3.5 fall significantly behind by about 8\%. This underscores the need for future efforts to enhance ToM capabilities on lighter-weight and open-source LLMs.

\paragraph{Ethics Statement}
The data utilized in our study is solely sourced from publicly available materials. No user information was involved or collected during our research. We are committed to upholding the principles of responsible and transparent data usage throughout our work.

Our dataset consists of several old movies that were created during a specific time period. As a result, they may exhibit certain limitations in terms of their content and representation, raising considerations of fairness. For instance, in the movie \emph{King Kong}, the heroine is portrayed as a blonde and beautiful lady, which may reflect certain biases of the era. Additionally, some movies in our dataset contain vulgar language and are classified as R-rated, e.g., \emph{South Park: Bigger, Longer \& Uncut}.

However, it's important to note that our dataset construction strategy can be used to create more instances with newly coming movie scripts. In the future, we plan to curate a subset of movies that are free from such biases and limitations. We currently include the existing list of movies simply to ensure that our dataset is sufficiently large. By doing so, as the first endeavor in this direction, we aim to draw statistically more accurate conclusions.

Lastly, as mentioned in the ``Limitations'' section, it is important to address the existing gender bias in the current dataset, where the majority of main characters are male. To mitigate this bias, future work needs to manually and carefully curate new evaluation sets that exhibit a balanced representation of character genders.

\bibliography{references,atoms}

\begin{thebibliography}{44}
\providecommand{\natexlab}[1]{#1}
\providecommand{\url}[1]{\texttt{#1}}
\expandafter\ifx\csname urlstyle\endcsname\relax
  \providecommand{\doi}[1]{doi: #1}\else
  \providecommand{\doi}{doi: \begingroup \urlstyle{rm}\Url}\fi

\bibitem[Bansal et~al.(2020)Bansal, Jha, and McCallum]{bansal2019learning}
Bansal, T., Jha, R., and McCallum, A.
\newblock Learning to few-shot learn across diverse natural language classification tasks.
\newblock In \emph{Proceedings of the 28th International Conference on Computational Linguistics}. International Committee on Computational Linguistics, 2020.

\bibitem[Brahman et~al.(2021)Brahman, Huang, Tafjord, Zhao, Sachan, and Chaturvedi]{brahman2021let}
Brahman, F., Huang, M., Tafjord, O., Zhao, C., Sachan, M., and Chaturvedi, S.
\newblock {``}let your characters tell their story{''}: A dataset for character-centric narrative understanding.
\newblock In \emph{Findings of the Association for Computational Linguistics: EMNLP 2021}. Association for Computational Linguistics, 2021.

\bibitem[Brown et~al.(2020)Brown, Mann, Ryder, Subbiah, Kaplan, Dhariwal, Neelakantan, Shyam, Sastry, Askell, et~al.]{brown2020language}
Brown, T., Mann, B., Ryder, N., Subbiah, M., Kaplan, J.~D., Dhariwal, P., Neelakantan, A., Shyam, P., Sastry, G., Askell, A., et~al.
\newblock Language models are few-shot learners.
\newblock \emph{Advances in neural information processing systems}, 33:\penalty0 1877--1901, 2020.

\bibitem[Bubeck et~al.(2023)Bubeck, Chandrasekaran, Eldan, Gehrke, Horvitz, Kamar, Lee, Lee, Li, Lundberg, et~al.]{bubeck2023sparks}
Bubeck, S., Chandrasekaran, V., Eldan, R., Gehrke, J., Horvitz, E., Kamar, E., Lee, P., Lee, Y.~T., Li, Y., Lundberg, S., et~al.
\newblock Sparks of artificial general intelligence: Early experiments with gpt-4.
\newblock \emph{arXiv preprint arXiv:2303.12712}, 2023.

\bibitem[Chase(2022)]{NC}
Chase, N.
\newblock {How to Introduce Characters in a Screenplay} neilchasefilm.
\newblock {https://neilchasefilm.com/how-to-introduce-characters-in-a-screenplay/}, 2022.
\newblock Accessed: 2022-05-07.

\bibitem[Chen et~al.(2017)Chen, Zhou, and Choi]{chen2017robust}
Chen, H.~Y., Zhou, E., and Choi, J.~D.
\newblock Robust coreference resolution and entity linking on dialogues: Character identification on tv show transcripts.
\newblock In \emph{Proceedings of CoNLL 2017}, pp.\  216--225, 2017.

\bibitem[Chen \& Choi(2016)Chen and Choi]{chen2016character}
Chen, Y.-H. and Choi, J.~D.
\newblock Character identification on multiparty conversation: Identifying mentions of characters in tv shows.
\newblock In \emph{Proceedings of SIGDIAL 2016}, pp.\  90--100, 2016.

\bibitem[Cohen(2021)]{cohen2021exploring}
Cohen, M.
\newblock Exploring roberta's theory of mind through textual entailment.
\newblock 2021.

\bibitem[Coucke et~al.(2018)Coucke, Saade, Ball, Bluche, Caulier, Leroy, Doumouro, Gisselbrecht, Caltagirone, Lavril, et~al.]{coucke2018snips}
Coucke, A., Saade, A., Ball, A., Bluche, T., Caulier, A., Leroy, D., Doumouro, C., Gisselbrecht, T., Caltagirone, F., Lavril, T., et~al.
\newblock Snips voice platform: an embedded spoken language understanding system for private-by-design voice interfaces.
\newblock \emph{arXiv preprint arXiv:1805.10190}, 2018.

\bibitem[Finn et~al.(2017)Finn, Abbeel, and Levine]{finn2017model}
Finn, C., Abbeel, P., and Levine, S.
\newblock Model-agnostic meta-learning for fast adaptation of deep networks.
\newblock In \emph{International conference on machine learning}, pp.\  1126--1135. PMLR, 2017.

\bibitem[Flekova \& Gurevych(2015)Flekova and Gurevych]{flekova2015personality}
Flekova, L. and Gurevych, I.
\newblock Personality profiling of fictional characters using sense-level links between lexical resources.
\newblock In \emph{Proceedings of the 2015 Conference on Empirical Methods in Natural Language Processing}, pp.\  1805--1816, 2015.

\bibitem[Gernsbacher et~al.(1998)Gernsbacher, Hallada, and Robertson]{gernsbacher1998automatically}
Gernsbacher, M.~A., Hallada, B.~M., and Robertson, R.~R.
\newblock How automatically do readers infer fictional characters' emotional states?
\newblock \emph{Scientific studies of reading}, 2\penalty0 (3):\penalty0 271--300, 1998.

\bibitem[Han et~al.(2018)Han, Zhu, Yu, Wang, Yao, Liu, and Sun]{han2018fewrel}
Han, X., Zhu, H., Yu, P., Wang, Z., Yao, Y., Liu, Z., and Sun, M.
\newblock {F}ew{R}el: A large-scale supervised few-shot relation classification dataset with state-of-the-art evaluation.
\newblock In \emph{Proceedings of the 2018 Conference on Empirical Methods in Natural Language Processing}. Association for Computational Linguistics, 2018.

\bibitem[He et~al.(2023)He, Wu, Jia, Mihalcea, Chen, and Deng]{he2023hi}
He, Y., Wu, Y., Jia, Y., Mihalcea, R., Chen, Y., and Deng, N.
\newblock Hi-tom: A benchmark for evaluating higher-order theory of mind reasoning in large language models.
\newblock In \emph{Findings of the Association for Computational Linguistics: EMNLP 2023}, 2023.

\bibitem[Hou et~al.(2020)Hou, Mao, Lai, Chen, Che, Chen, and Liu]{hou2020fewjoint}
Hou, Y., Mao, J., Lai, Y., Chen, C., Che, W., Chen, Z., and Liu, T.
\newblock Fewjoint: a few-shot learning benchmark for joint language understanding.
\newblock \emph{arXiv preprint arXiv:2009.08138}, 2020.

\bibitem[Jahan et~al.(2021)Jahan, Mittal, and Finlayson]{jahan2021inducing}
Jahan, L., Mittal, R., and Finlayson, M.
\newblock Inducing stereotypical character roles from plot structure.
\newblock In \emph{Proceedings of the 2021 Conference on Empirical Methods in Natural Language Processing}, pp.\  492--497, 2021.

\bibitem[Keysar et~al.(2000)Keysar, Barr, Balin, and Brauner]{keysar2000taking}
Keysar, B., Barr, D.~J., Balin, J.~A., and Brauner, J.~S.
\newblock Taking perspective in conversation: The role of mutual knowledge in comprehension.
\newblock \emph{Psychological Science}, 11\penalty0 (1):\penalty0 32--38, 2000.

\bibitem[Kosinski(2023)]{kosinski2023theory}
Kosinski, M.
\newblock Theory of mind may have spontaneously emerged in large language models.
\newblock \emph{arXiv preprint arXiv:2302.02083}, 2023.

\bibitem[Larson et~al.(2019)Larson, Mahendran, Peper, Clarke, Lee, Hill, Kummerfeld, Leach, Laurenzano, Tang, and Mars]{larson2019evaluation}
Larson, S., Mahendran, A., Peper, J.~J., Clarke, C., Lee, A., Hill, P., Kummerfeld, J.~K., Leach, K., Laurenzano, M.~A., Tang, L., and Mars, J.
\newblock An evaluation dataset for intent classification and out-of-scope prediction.
\newblock In \emph{Proceedings of the 2019 Conference on Empirical Methods in Natural Language Processing and the 9th International Joint Conference on Natural Language Processing (EMNLP-IJCNLP)}. Association for Computational Linguistics, 2019.

\bibitem[Li et~al.(2020)Li, Jiang, Feng, Sprague, Zhou, and Hoey]{li2020aloha}
Li, A.~W., Jiang, V., Feng, S.~Y., Sprague, J., Zhou, W., and Hoey, J.
\newblock Aloha: Artificial learning of human attributes for dialogue agents.
\newblock In \emph{Proceedings of the AAAI Conference on Artificial Intelligence}, volume~34, pp.\  8155--8163, 2020.

\bibitem[Ma et~al.(2011)Ma, Vandekerckhove, Van~Overwalle, Seurinck, and Fias]{ma2011spontaneous}
Ma, N., Vandekerckhove, M., Van~Overwalle, F., Seurinck, R., and Fias, W.
\newblock Spontaneous and intentional trait inferences recruit a common mentalizing network to a different degree: spontaneous inferences activate only its core areas.
\newblock \emph{Social neuroscience}, 6\penalty0 (2):\penalty0 123--138, 2011.

\bibitem[Ma et~al.(2023)Ma, Sansom, Peng, and Chai]{ma2023towards}
Ma, Z., Sansom, J., Peng, R., and Chai, J.
\newblock Towards a holistic landscape of situated theory of mind in large language models.
\newblock In \emph{Findings of the Association for Computational Linguistics: EMNLP 2023}, pp.\  1011--1031, 2023.

\bibitem[Massey et~al.(2015)Massey, Xia, Bamman, and Smith]{massey2015annotating}
Massey, P., Xia, P., Bamman, D., and Smith, N.~A.
\newblock Annotating character relationships in literary texts.
\newblock \emph{arXiv:1512.00728}, 2015.

\bibitem[Nematzadeh et~al.(2018)Nematzadeh, Burns, Grant, Gopnik, and Griffiths]{nematzadeh2018evaluating}
Nematzadeh, A., Burns, K., Grant, E., Gopnik, A., and Griffiths, T.
\newblock Evaluating theory of mind in question answering.
\newblock In \emph{Proceedings of the 2018 Conference on Empirical Methods in Natural Language Processing}. Association for Computational Linguistics, 2018.

\bibitem[Perner \& Wimmer(1985)Perner and Wimmer]{perner1985john}
Perner, J. and Wimmer, H.
\newblock “john thinks that mary thinks that…” attribution of second-order beliefs by 5-to 10-year-old children.
\newblock \emph{Journal of experimental child psychology}, 39\penalty0 (3):\penalty0 437--471, 1985.

\bibitem[Premack \& Woodruff(1978)Premack and Woodruff]{premack1978does}
Premack, D. and Woodruff, G.
\newblock Does the chimpanzee have a theory of mind?
\newblock \emph{Behavioral and brain sciences}, 1\penalty0 (4):\penalty0 515--526, 1978.

\bibitem[Riley(2009)]{riley2009hollywood}
Riley, C.
\newblock \emph{The Hollywood standard: the complete and authoritative guide to script format and style}.
\newblock Michael Wiese Productions, 2009.

\bibitem[Rowe et~al.(2008)Rowe, Ha, and Lester]{rowe2008archetype}
Rowe, J.~P., Ha, E.~Y., and Lester, J.~C.
\newblock Archetype-driven character dialogue generation for interactive narrative.
\newblock In \emph{International Workshop on Intelligent Virtual Agents}, pp.\  45--58. Springer, 2008.

\bibitem[Sang et~al.(2022{\natexlab{a}})Sang, Mou, Yu, Wang, Li, and Stanton]{sang2022mbti}
Sang, Y., Mou, X., Yu, M., Wang, D., Li, J., and Stanton, J.
\newblock Mbti personality prediction for fictional characters using movie scripts.
\newblock \emph{arXiv preprint arXiv:2210.10994}, 2022{\natexlab{a}}.

\bibitem[Sang et~al.(2022{\natexlab{b}})Sang, Mou, Yu, Yao, Li, and Stanton]{sang2022tvshowguess}
Sang, Y., Mou, X., Yu, M., Yao, S., Li, J., and Stanton, J.
\newblock Tvshowguess: Character comprehension in stories as speaker guessing.
\newblock \emph{arXiv preprint arXiv:2204.07721}, 2022{\natexlab{b}}.

\bibitem[Sap et~al.(2019)Sap, Rashkin, Chen, Le~Bras, and Choi]{sap2019social}
Sap, M., Rashkin, H., Chen, D., Le~Bras, R., and Choi, Y.
\newblock Social {IQ}a: Commonsense reasoning about social interactions.
\newblock In \emph{Proceedings of the 2019 Conference on Empirical Methods in Natural Language Processing and the 9th International Joint Conference on Natural Language Processing (EMNLP-IJCNLP)}, pp.\  4463--4473, Hong Kong, China, November 2019. Association for Computational Linguistics.

\bibitem[Shapira et~al.(2023)Shapira, Levy, Alavi, Zhou, Choi, Goldberg, Sap, and Shwartz]{shapira2023clever}
Shapira, N., Levy, M., Alavi, S.~H., Zhou, X., Choi, Y., Goldberg, Y., Sap, M., and Shwartz, V.
\newblock Clever hans or neural theory of mind? stress testing social reasoning in large language models, 2023.

\bibitem[Sileo \& Lernould(2023)Sileo and Lernould]{sileo2023mindgames}
Sileo, D. and Lernould, A.
\newblock Mindgames: Targeting theory of mind in large language models with dynamic epistemic modal logic.
\newblock \emph{arXiv preprint arXiv:2305.03353}, 2023.

\bibitem[Snell et~al.(2017)Snell, Swersky, and Zemel]{snell2017prototypical}
Snell, J., Swersky, K., and Zemel, R.
\newblock Prototypical networks for few-shot learning.
\newblock \emph{Advances in neural information processing systems}, 30, 2017.

\bibitem[Tracey et~al.(2022)Tracey, Rambow, Cardie, Dalton, Dang, Diab, Dorr, Guthrie, Markowska, Muresan, et~al.]{tracey2022best}
Tracey, J., Rambow, O., Cardie, C., Dalton, A., Dang, H.~T., Diab, M., Dorr, B., Guthrie, L., Markowska, M., Muresan, S., et~al.
\newblock Best: The belief and sentiment corpus.
\newblock In \emph{Proceedings of the Thirteenth Language Resources and Evaluation Conference}, pp.\  2460--2467, 2022.

\bibitem[Vinyals et~al.(2016)Vinyals, Blundell, Lillicrap, Wierstra, et~al.]{vinyals2016matching}
Vinyals, O., Blundell, C., Lillicrap, T., Wierstra, D., et~al.
\newblock Matching networks for one shot learning.
\newblock \emph{Advances in neural information processing systems}, 29, 2016.

\bibitem[Wolf et~al.(2020)Wolf, Debut, Sanh, Chaumond, Delangue, Moi, Cistac, Rault, Louf, Funtowicz, Davison, Shleifer, von Platen, Ma, Jernite, Plu, Xu, Le~Scao, Gugger, Drame, Lhoest, and Rush]{wolf-etal-2020-transformers}
Wolf, T., Debut, L., Sanh, V., Chaumond, J., Delangue, C., Moi, A., Cistac, P., Rault, T., Louf, R., Funtowicz, M., Davison, J., Shleifer, S., von Platen, P., Ma, C., Jernite, Y., Plu, J., Xu, C., Le~Scao, T., Gugger, S., Drame, M., Lhoest, Q., and Rush, A.
\newblock Transformers: State-of-the-art natural language processing.
\newblock In \emph{Proceedings of the 2020 Conference on Empirical Methods in Natural Language Processing: System Demonstrations}. Association for Computational Linguistics, 2020.

\bibitem[Wu et~al.(2023)Wu, Chen, Deng, Sabour, and Huang]{wu2023coke}
Wu, J., Chen, Z., Deng, J., Sabour, S., and Huang, M.
\newblock Coke: A cognitive knowledge graph for machine theory of mind.
\newblock \emph{arXiv preprint arXiv:2305.05390}, 2023.

\bibitem[Ye et~al.(2021)Ye, Lin, and Ren]{ye2021crossfit}
Ye, Q., Lin, B.~Y., and Ren, X.
\newblock {C}ross{F}it: A few-shot learning challenge for cross-task generalization in {NLP}.
\newblock In \emph{Proceedings of the 2021 Conference on Empirical Methods in Natural Language Processing}. Association for Computational Linguistics, 2021.

\bibitem[Yu et~al.(2018)Yu, Guo, Yi, Chang, Potdar, Cheng, Tesauro, Wang, and Zhou]{yu2018diverse}
Yu, M., Guo, X., Yi, J., Chang, S., Potdar, S., Cheng, Y., Tesauro, G., Wang, H., and Zhou, B.
\newblock Diverse few-shot text classification with multiple metrics.
\newblock In \emph{Proceedings of the 2018 Conference of the North {A}merican Chapter of the Association for Computational Linguistics: Human Language Technologies, Volume 1 (Long Papers)}. Association for Computational Linguistics, 2018.

\bibitem[Yuan et~al.(2020)Yuan, Fu, Shen, Xu, Shen, and Zhu]{yuan2020emergence}
Yuan, L., Fu, Z., Shen, J., Xu, L., Shen, J., and Zhu, S.-C.
\newblock Emergence of pragmatics from referential game between theory of mind agents.
\newblock \emph{arXiv preprint arXiv:2001.07752}, 2020.

\bibitem[Zhang \& Chai(2010)Zhang and Chai]{zhang2010towards}
Zhang, C. and Chai, J.~Y.
\newblock Towards conversation entailment: An empirical investigation.
\newblock In \emph{Proceedings of the 2010 Conference on Empirical Methods in Natural Language Processing}, pp.\  756--766, 2010.

\bibitem[Zhou et~al.(2023)Zhou, Zhu, Hu, Pujara, Ren, Callison-Burch, Choi, and Ammanabrolu]{zhou2023i}
Zhou, P., Zhu, A., Hu, J., Pujara, J., Ren, X., Callison-Burch, C., Choi, Y., and Ammanabrolu, P.
\newblock I cast detect thoughts: Learning to converse and guide with intents and theory-of-mind in dungeons and dragons.
\newblock In \emph{Proceedings of the 61th Annual Meeting of the Association for Computational Linguistics}, 2023.

\bibitem[Zhu et~al.(2021)Zhu, Neubig, and Bisk]{zhu2021few}
Zhu, H., Neubig, G., and Bisk, Y.
\newblock Few-shot language coordination by modeling theory of mind.
\newblock In \emph{International Conference on Machine Learning}, pp.\  12901--12911. PMLR, 2021.

\end{thebibliography}
\bibliographystyle{foobar2024}

\newpage
\appendix
\onecolumn
\section{Perturbation Setting}
\label{app:perturbation_setting}
To address the memorization problem of LLMs, 
we substitute the character names correspondingly with common names in Table \ref{tab:name_map} based on their gender as the \emph{perturbation setting}.
Specifically, for each movie, we first construct the mapping between the real name and the perturbed name for each character, then use this mapping to perform the perturbation throughout the movie.
We employ \texttt{gender-guesser}\footnote{\url{https://pypi.org/project/gender-guesser/}} to identify the gender of character names.
We adopt perturbation in the training scenes and the guessing options in the anonymized testing scenes for ToMPro, and the guessing options of both few-shot examples and test cases for ICL.
Additionally, movie titles are also redacted in ICL.

\begin{table}[H]
    \centering
    \caption{Common names for perturbation.}
    \vspace{-0.7\baselineskip}
    \begin{tabular}{l lllll}
    \toprule
     \bf Male   & David & James & John & Michael & Robert \\
    \midrule
     \bf Female & Elizabeth & Emily & Jennifer & Linda & Mary \\
    \midrule
     \bf Androgynous & Alex & Casey & Jordan & Morgan & Taylor \\
    \bottomrule
    \end{tabular}
    \label{tab:name_map}
\end{table}

\section{Details of the Implementations of the Non-LLM Baselines}
\subsection{Details of the Base Learner Architecture}
\label{app:base_learner}

We follow \cite{sang2022tvshowguess} to use the longformer-based character predictor (\textbf{Longformer-P}) as our character encoder, as shown in Figure~\ref{fig:model} (top).
This architecture consists of two steps: encoding the scene into contextualized embeddings and then conducting attentive pooling to obtain character representations in the scene.

(1) \emph{Scene Encoding:}
The input $S=T_0 \oplus T_1 \oplus ... \oplus T_N$ to the model is the concatenation of all the utterances $T_i$s in an anonymous scene.
Each $T_i$ has its text $U_i$ prefixed by a speaker ID token $\text{[P}_{x_i}\text{]}$ and suffixed by a separation $\text{[SPLIT]}$ token, \emph{i.e.},
\begin{equation}
\small
\setlength{\abovedisplayskip}{8pt}
\setlength{\belowdisplayskip}{4pt}
\begin{aligned}
    T_i = \text{[P}_{x_i}\text{]} \oplus U_i \oplus \text{[SPLIT]} \label{eq:utterance}   
\end{aligned}
\end{equation}
where $\text{P}_{x_i}\in{\text{P}_0, \text{P}_1, \cdots}$. 
We use a Longformer to encode the whole input $S$ to get its contextualized embedding, \emph{i.e.}, $\mathbf{H} = \textrm{Longformer}(S) \in \mathbb{R}^{L \times D}$.

(2) \emph{Attentive Pooling per Character:}
For each character ID $\text{P}_{x}$, we introduce a mask $M_x \in \mathbb{R}^{L \times 1}$, such that $M_x[j]=1$ if the $j$-th word belongs to an utterance of $\text{P}_{x}$; and 0 otherwise.
For each character $\text{P}_{x}$, we then collect the useful information from all their utterances as masked by $M_x$ as
\begin{equation}
\small
\setlength{\abovedisplayskip}{8pt}
\setlength{\belowdisplayskip}{4pt}
\begin{aligned}
    A &= \textrm{Attention}(\mathbf{H}), \, \alpha_x = \textrm{Softmax}(A \odot M_x). \nonumber
\end{aligned}
\end{equation}
The character-specific attention $\alpha_x$ is then used to pool the hidden states to summarize a character representation in the input scene ${S}$, $\mathbf{e}_{\text{P}_x|S} = \mathbf{H}^T \alpha_x$. 

\subsection{Details of the Prototypical Network Implementation}
\label{app:proto_net}
When applying the Prototypical Network~\cite{snell2017prototypical} to our \datasetnamens, each task $T_i$ corresponds to a movie $M_i$.
For each masked character $\text{P}_x$ in a training scene $S \in M_i$, by applying a base learner as the metric network $\Lambda(\cdot)$, we achieve the character embedding conditioned on the scene as $\mathbf{e}_{\text{P}_x|S}=\Lambda(\text{P}_x, S)$. Then for each main character $c_i$, we compute the prototype as $\mathbf{e}_{c_k} = \textrm{avg}\left(\left \{\mathbf{e}_{\text{P}_x|S}| (\text{P}_x=c_k, S) \in \mathcal{D}^{train}_i \right \}\right)$.
For a testing case $x'$ that corresponds to a $\text{P}_x' \in S'$, the prediction logits of $\textrm{score}(\text{P}_x'=c_k)=\textrm{cos}(\mathbf{e}_{\text{P}_x'|S'}, \mathbf{e}_{c_k})$.
The whole inference process is shown in Figure~\ref{fig:model} (middle). 

During training, because some of the movies have a large number ($>$100) of training scenes, computing the prototypes from the full training scenes for each training iteration is time-consuming. We sample at most 5 support mini-batches (8 scenes in each batch) to compute the prototypes.
Updating all the support instances together with the training instances also leads to memory issues; we fix the support embedding branch to overcome this issue.

\subsection{Details of the LEOPARD Implementation}
\label{app:leopard}

To handle the challenge of varying numbers of classes across tasks, the LEOPARD introduce an additional parameter generator to MAML, which learns to generate the initial parameters of the prediction layer for a new task. 
We adapted LEOPARD to our problem as follows.
We denote the training set of a task $T_i \in \mathcal{T}^{Train}$ as $\mathcal{D}^{train}_i$.
First, we sample a few scenes $\mathcal{S}^{(i)}_k=\{(\text{P}_x=c_k, S)\} \in \mathcal{D}^{train}_i$ as the support set for each character $c_k$ and compute its character embedding in the masked scene $S$ as:
$\mathbf{e}_{c_k|S}\doteq\mathbf{e}_{\text{P}_x|S}=\Lambda_\theta\left(\text{P}_x, S\right)$, where $\Lambda_\theta(\cdot)$ can be initialized with a trained prototypical network.

Second, LEOPARD generates a linear model ($\mathbf{w}^{(i)}_k$, $b^{(i)}_k$) for each class $c^{(i)}_k$, as task $i$'s prediction layer:
\begin{equation}
\small
\setlength{\abovedisplayskip}{8pt}
\setlength{\belowdisplayskip}{4pt}
\begin{aligned}
    \mathbf{w}^{(i)}_k, b^{(i)}_k = {\sum_{S \in \mathcal{S}^{(i)}_k}g_\psi\left(\mathbf{e}_{c_k|S} \right)}/{\left|\mathcal{S}^{(i)}_k\right|}, \nonumber
\end{aligned}
\end{equation}
where $g_\psi$ is an MLP with two layers and $\tanh$ activation.
Then, for task $T_i$, we obtain its weight matrix $\mathbf{W}^{(i)}$ and bias $\mathbf{b}^{(i)}$ in the prediction layer by concatenating the weights and bias of all classes:
\begin{equation*}
\small
\setlength{\abovedisplayskip}{8pt}
\setlength{\belowdisplayskip}{4pt}
    \begin{aligned}
        \mathbf{W}^{(i)} = \left[\mathbf{w}^{(i)}_1; \cdots; \mathbf{w}^{(i)}_{N_i}\right]\ \ \mathbf{b}^{(i)} = \left[b^{(i)}_1; \cdots; b^{(i)}_{N_i}\right].
    \end{aligned}
\end{equation*}

Finally, given this meta-predicted layer, the prediction given an input $(\text{P}_x, S)$ can be obtained by
\begin{equation*}
\small
\setlength{\abovedisplayskip}{8pt}
\setlength{\belowdisplayskip}{4pt}
    \begin{aligned}
        p\left(\text{P}_x=c | S\right) = \mathrm{softmax}(\mathbf{W}^{(i)} h_\phi\left(\Lambda_\theta\left(\text{P}_x, S\right)\right) + \mathbf{b}^{(i)})
    \end{aligned}
\end{equation*}
where $h_\phi$ is another MLP with parameters $\phi$ to map the instance embedding to the $l$-dimensional space.
\newpage

\section{Details of In-Context Learning Solution with LLMs}
\label{app:llm_prompt}
We demonstrate our prompt template in Figure~\ref{app:few_shot_template}. $k$ examples from the training scenes are demonstrated at the beginning of the prompt together with the correct answers.
\begin{figure}[H]
    \centering
\begin{lstlisting}[language=prompt]
<<<[SYSTEM]>>>
You are playing a game guessing the identity of anonymized characters from movie scenes. For each testing case, a few history scenes from the same movie are provided as examples and background.

<<<[USER]>>>
// Example 1
Please read the following scene from the movie "[[[MOVIE_NAME]]]":

{{ scene title }}
P0: {{ utterance }}
{{ narration }}
P1: {{ utterance }}
P0: {{ utterance }}
P1: {{ utterance }}
{{ ... the rest of the scene is omitted ... }}

Choose who the anonymous characters (===P0, P1===) are from the follows:
(a) [[[CHAR1]]]
(b) [[[CHAR2]]]
(c) [[[CHAR3]]]
(d) [[[CHAR4]]]
(e) [[[CHAR5]]]

Please answer in a format like: ===P[0-1]===-{[[[CHAR1]]], [[[CHAR2]]], [[[CHAR3]]], [[[CHAR4]]], [[[CHAR5]]]}.
Answer:
===P0===-[[[CHAR4]]]
===P1===-[[[CHAR2]]]

// Example 2 - k
{{ ... examples are omitted ... }}

// Test Case
Please read the following scene from the movie "[[[MOVIE_NAME]]]":

{{ scene title }}
P0: {{ utterance }}
P1: {{ utterance }}
P2: {{ utterance }}
{{ ... the rest of the scene is omitted ... }}

Choose who the anonymous characters (===P0, P1, P2===) are from the follows:
(a) [[[CHAR1]]]
(b) [[[CHAR2]]]
(c) [[[CHAR3]]]
(d) [[[CHAR4]]]
(e) [[[CHAR5]]]

Please answer in a format like: ===P[0-2]===-{[[[CHAR1]]], [[[CHAR2]]], [[[CHAR3]]], [[[CHAR4]]], [[[CHAR5]]]}.
\end{lstlisting}
    \caption{GPT-4 Prompt Template with Few-Shot Enhancement}
    \label{app:few_shot_template}
\end{figure}

\newpage

\section{Details of ToM Prompting Solution with LLMs}
\label{app:tompro}

In the long text input setting, we suggest a recurrent method to update character modeling by processing scenes in chronological order with LLMs.
As depicted in Figure \ref{fig:llm-methods}, our approach divides character modeling into personalities and the other 4 instant theory-of-mind dimensions, \textit{i.e.}, emotions, beliefs, desires, and intentions.
Personalities $P^{(t)}$ are modeled using a recurrent memory that updates chronologically through scene iterations, while the other 4 instant dimensions $I^{(t)}$ are all derived from the current scene.
\begin{align}
P^{(t)} &= f_{\text{personalities}}([S^{(t)}, \{\text{P}_x = c_k\}^{(t)}], P^{(t-1)})   \\
I^{(t)} &= f_{\text{instant dims}}([S^{(t)}, \{\text{P}_x = c_k\}^{(t)}])
\end{align}
In the testing phase, we create a guessing prompt that incorporates character modeling obtained in the previous steps attached to the choices.
\begin{equation}
\{\text{P}_x = c_k\}^{(t)} = f_{\text{guess}}(S^{(t)}, P^{(t)}, I^{(t)}), \quad S^{(t)}\in\mathcal{D}^{test}
\end{equation}
Complete prompt templates for $f_{\text{personalities}}$ and $f_{\text{instant dims}}$ are provided in Figure \ref{app:tompro-tom} and \ref{app:tompro-personality}. The guessing prompt $f_{\text{guess}}$ is provided in Figure \ref{fig:llm-methods}b.

\begin{figure}[H]
    \centering
\begin{lstlisting}[language=prompt]
<<<[SYSTEM]>>>
You are expected to use theory of mind to elucidate the |||emotions, beliefs, desires, intentions||| of [[[CHAR]]] based on the dialogue within a scene.

<<<[USER]>>>
You are expected to use theory of mind to elucidate the |||emotions, beliefs, desires, intentions||| of [[[CHAR]]] based on the dialogue within a scene.

The following are explanations of |||emotions, beliefs, desires, intentions|||:
|||Emotions|||: Emotions are strong feelings deriving from one's circumstances, mood, or relationships with others. And emotions are variously associated with thoughts, feelings, behavioral responses, and a degree of pleasure or displeasure.
|||Beliefs|||: Beliefs encompass both objective facts and subjective perceptions concerning the existence or truth of something.
|||Desires|||: Desires encompass both physical needs and psychological yearnings. Desires incline people toward action and fulfilling desires is pleasurable.
|||Intentions|||: Intentions are blueprints that steer actions, encompassing both future plans and the motivations driving current behavior.

Please read the following scene from a movie:
|||[Scene]|||
[[[{{ original scene with character names }}]]]

Please use theory of mind to elucidate the |||emotions, beliefs, desires, intentions||| of [[[CHAR]]]. Please answer in a format like: "Emotions:\nBeliefs:\nDesires:\nIntentions:".
\end{lstlisting}
    \caption{Prompt template for 4 instant ToM dimensions (emotions, beliefs, desires, and intentions).}
    \label{app:tompro-tom}
\end{figure}
\newpage
\begin{figure}[H]
    \centering
\begin{lstlisting}[language=prompt]
<<<[SYSTEM]>>>
We're going to play a game to update the given character descriptions based on a new scene from a movie.

<<<[USER]>>>
We're going to play a game to update the given character descriptions based on a new scene from a movie. First, I will give you the character descriptions of each character. So you will learn about the characteristics of these characters. Then, I will give you a scene so that you will have a new understanding of each character. What you have to do is to improve the character descriptions of the character based on the original character descriptions and the unique traits exhibited by the characters in the scene. Be careful not to change the original character descriptions easily, because you need to control the proportion of a scene's understanding of the character. And the format of the new character descriptions should be like the given character descriptions.

You are given the following character descriptions of some characters:
|||[Character Descriptions]|||
[[[CHAR1]]]: [[[DESC1]]]
[[[CHAR2]]]: [[[DESC2]]]
[[[CHAR3]]]: [[[DESC3]]]
[[[CHAR4]]]: [[[DESC4]]]

Please read the following scene from a movie about some characters ([[[CHAR2]]], [[[CHAR3]]]):
|||[Scene]|||
[[[{{ original scene with character names }}]]]

Now you can update and improve the character description for each character ([[[CHAR2]]], [[[CHAR3]]]) by incorporating the unique traits exhibited by the characters in the scene. Please accentuate the unique personality traits, specific interests or hobbies, unique backgrounds or experiences of each character, while reducing the homogeneity among the six characters, thus creating unique and individualized character descriptions for each. Please provide an objective character description of each character, encompassing both positive and negative personality traits, with particular attention to not overlooking any character weaknesses. Please explicitly mention the characters' weaknesses. And don't provide suggestions regarding their weaknesses. Please don't delete the unique traits in the original character descriptions. Please preserve their most important and unique personality traits. If a character's character description doesn't need to be adjusted, just keep it as it is. And be careful the specific scene content should not appear in the character descriptions, because the character descriptions is a brief generalization of a character and not a generalization of the scene. You don't need to explain the reasons for adjustments, just provide the character descriptions. When the character dscriptions need to be compressed due to excessive length, please retain the unique characteristics of each character, and abbreviate or disregard similar or identical features. Please answer in a format like: "[[[CHAR2]]]: [[[CHAR2]]] ...\n[[[CHAR3]]]: [[[CHAR3]]]". And do not answer other than that.
\end{lstlisting}
    \caption{Prompt template for personalities.}
    \label{app:tompro-personality}
\end{figure}
\newpage
\section{Human Annotation}

\subsection{Movies Used for Human Study}
\label{app:human_annotation}
Two raters evaluated 11 movies from movie genre that have more than 100 movies. Table~\ref{tab:dev_sampled_movies} shows the movie names used in human study.

\subsection{Interface for the Human Study}
\label{app:interface}
Figure \ref{fig: human_study_interface} shows the interfaces of the human study.

\subsection{Annotator Accuracy}
\label{app:annotation accuracy}

We provide the breakdown of the two annotators' accuracy in Table~\ref{tab:annAcc1} and \ref{tab:annAcc2} for in-depth comparison to the models and for understanding the ratio of testing examples that require history information.

\subsection{Examples of Human Errors}
\label{app:human_error}
Table \ref{tab: mistake} provides an example of human mistake cases and Table \ref{tab: unsolvable} provides an example of unsolvable cases. The human mislabeled characters are marked as red.

\subsection{Remark on Human Solutions that Related to Meta-Learning of ToM}
\label{app:human_solution}
Our human study revealed that to solve our task, humans frequently leverage their knowledge from seen movies, which corresponds to a ``meta-learning'' style solution.

{Specifically, in our human study, the raters reported the following strategies they used to understand a new character:}

(1) They first classify the characters to rough \textbf{archetypes} they learned from previous experience, \emph{e.g.}, \emph{Hero} and \emph{Villain} that are common in action movies. When our human raters have these concepts learned from their previous experience, they can quickly assign these coarse tags to the characters.

(2) When archetypes are insufficient, \emph{e.g.}, many characters are unconventional to the archetypes or when there are multiple characters with the same archetype in one movie,
they \textbf{associate} the new characters with the ones in the movies they have seen before, to make a fine-grained understanding.
For example, when labeling \emph{Tomorrow Never Ends}, the raters leverage their understanding of \emph{Ethan Hunt} in the movie \emph{Mission Impossible} to have a pre-impression of \emph{James Bond}.
Similarly, in the movie \emph{Ghost Ship}, the protagonist \emph{Epps} was the only person that survived. 
When evaluating the scene that a character saw the phantom girl, the rater intuitively considered \emph{Epps} might be the top candidate compared to other candidates.
because such gifts of seeing supernatural figures that other characters could not usually happen to ``\emph{heroes}'' in horror movies they have seen, such as \emph{Danny} in \emph{The Shinning} or \emph{Cole} in \emph{The Sixth Sense}.

\newpage
\begin{figure}[H]
\input{img/user-study}
\end{figure}

\newpage
\begin{table}[H]
\caption{Human study examples from development set.}
\vspace{-0.7\baselineskip}
\centering
\begin{tabular}{llcc} 
\toprule                
\bf Movie Genre &
\bf Example     &
\bf \#Scene       &
\bf \#Instances    \\
\midrule
 Action    & \emph{Rush Hour 2}, \emph{Aliens} & 88 & 143\\
 Adventure & \emph{Mission Impossible II, }\emph{Tomorrow Never Dies} & 41 & 61\\
 Comedy    & \emph{South Park} & 28 & 51\\
 Crime     & \emph{Croupier} & 46 & 60\\
 Drama     & \emph{King Kong} & 39 & 61\\
 Horror    & \emph{Ghost Ship} & 30 & 61\\
 Romance   & \emph{Last Tango in Paris} & 16 & 25\\
 Sci-Fi    & \emph{Jurassic Park the Lost World} & 23 & 60\\
 Thriller  & \emph{Very Bad Things} & 24 & 50\\
\bottomrule
\end{tabular}
\label{tab:dev_sampled_movies}
\end{table}

\begin{table}[H]
\centering
\caption{Annotator1 accuracy breakdown.}
\vspace{-0.7\baselineskip}
\begin{tabular}{lcccc} 
\toprule                
\bf Movie Name& \bf Correct& \bf \#Instances & \bf Requiring History& \bf \#Scenes\\\midrule
 \emph{Rush Hour}                     & 33  & 37  & 21  & 22  \\
 \emph{Very Bad Things}               & 22  & 27  & 12  & 12  \\
 \emph{Aliens}                        & 39  & 44  & 20  & 22  \\
 \emph{Last Tango in Paris}           & 11  & 11  & 8   & 8   \\
 \emph{Croupier}                      & 29  & 33  & 21  & 23  \\
 \emph{South Park}                    & 22  & 27  & 14  & 14  \\
 \emph{Mission Impossible II}         & 7   & 8   & 3   & 4   \\
 \emph{Tomorrow Never Dies}           & 18  & 18  & 15  & 16  \\
 \emph{Jurassic Park the Lost World}  & 28  & 33  & 10  & 11  \\
 \emph{Ghost Ship}                    & 24  & 31  & 14  & 15  \\
 \emph{King Kong}                     & 27  & 29  & 16  & 19  \\
\midrule
 \bf Total                            & 260 & 298 & 154 & 166 \\
\bottomrule
\end{tabular}
\label{tab:annAcc1}
\end{table}

\begin{table}[H]
\centering
\caption{Annotator2 accuracy breakdown.}
\vspace{-0.7\baselineskip}
\begin{tabular}{lcccc} 
\toprule                
\bf Movie Name& \bf Correct& \bf \#Instances & \bf Requiring History& \bf \#Scenes\\\midrule
 \emph{Rush Hour}                     & 24  & 29  & 21  & 22  \\
 \emph{Very Bad Things}               & 23  & 23  & 11  & 12  \\
 \emph{Aliens}                        & 27  & 33  & 21  & 22  \\
 \emph{Last Tango in Paris}           & 10  & 14  & 7   & 8   \\
 \emph{Croupier}                      & 27  & 27  & 21  & 23  \\
 \emph{South Park}                    & 22  & 24  & 14  & 14  \\
 \emph{Mission Impossible II}         & 14  & 14  & 4   & 5   \\
 \emph{Tomorrow Never Dies}           & 21  & 21  & 15  & 16  \\
 \emph{Jurassic Park the Lost World}  & 25  & 27  & 12  & 12  \\
 \emph{Ghost Ship}                    & 21  & 27  & 13  & 15  \\
 \emph{King Kong}                     & 29  & 32  & 18  & 20  \\
\midrule
 \bf Total                            & 243 & 271 & 157 & 169 \\
\bottomrule
\end{tabular}
\label{tab:annAcc2}
\end{table}

\newpage
\begin{figure}[H]
\centering
\begin{lstlisting}[language=ioexample]
<<<[Human Mistake]>>>
:::Movie Name:::: ---Ghost Ship---
:::Background:::: ---[INT. Chimera - Aquarium tank - Later - Day]---
:::Candidates:::: ---{Epps, Greer, Dodge, Murphy}---
:::Background:::: ---[Greer sits in the tank, visible through a large piece of armored aquarium glass, amidst the fake coral. He sits in the sand hugging his legs to his chest, bobbing slightly as he speaks in his nonsensical language.]---
:::P0:::: Must've been him all along.
:::Background:::: ---[P0 and P1 look on from the outside in the promenade.]---
:::P0:::: Smashed the radio. Scuttled the boat. Killed dodge. Would've killed you. He's off his nut, no doubt there. 
:::Background:::: ---[They watch him in silence a moment as Greer mutters and bobs.]---
:::P0:::: What do you think? 
:::P1:::: Could be a stroke. Who knows? (a beat) The general log said the crew were fighting among themselves. "Like wild dogs." 
:::P0:::: Over the gold. 
:::P1:::: Maybe it was more than that.
:::Background:::: ---[Greer gets up, comes to the window, looking out at them. He presses his face to the glass.]---
:::P1:::: They went crazy.
:::P0:::: Crazy with greed. Not crazy. Not like him.
:::Background:::: ---[A beat as P1 looks off. In the window Greer drags his hideously distorted face over the glass, the blood from his wounds smearing in broad red streaks.]---
:::Answer:::: <<<P0>>>: Murphy, <<<P1>>>: Epps
\end{lstlisting}
\caption{Example of a human mistake.}
\label{tab: mistake}
\end{figure}

\begin{figure}[H]
\centering
\begin{lstlisting}[language=ioexample]
<<<[Unsolvable Case]>>>
:::Movie name:::: ---South Park---
:::Candidates:::: ---{Stan, Kyle, Cartman}---
:::Background:::: ---[EXT. inside the prison camp Mole pops his head out of the ground. immediately, a search light passes over the hole.]---
:::Background:::: ---[A beat... Then mole takes a long drag off his cigarette and slowly blows the smoke.]---
:::the Mole:::: Now listen carefully. Stan and Kyle, you stand watch here and await my return. if any guards come by, make a sound like a dying giraffe.
:::P0:::: What's a dying giraffe sound like?
:::the Mole:::: (Putting his hands to his mouth) Gwpaapa. Gwpaapa.
:::P0:::: Kay.
:::Background:::: ---[The Mole turns to P2.]---
:::the Mole:::: Cartman, over zere, is the electrical box. You must sneak over zere and shut it off before I return with terrance and phillip or the alarms will sound and i will be shot full of holes. got it?
:::P2:::: Okay.
:::the Mole:::: You must shut off the power, this is very important do you understa-
:::P2:::: I heard you the first time! I'm not Lou Ferigno for Pete's sake!
:::Background:::: ---[P2 storms off.]---
:::the Mole:::: I will tunnel my way into ze buildings, and find ze prisoners.
:::Background:::: ---[The mole starts to dig.]---
:::P0:::: Be careful, dude.
:::the Mole:::: Careful? Was my mother careful when she stabbed me in the heart with a clothes hanger while i was still in ze womb?
:::Background:::: ---[And with that, the mole quickly starts to tunnel his way underground.]---
:::P1:::: ---[Damn, dude, that kid is fucked up.]---
:::Answer:::: <<<P0>>>: Kyle, <<<P1>>>: Stan, P2: Cartman
\end{lstlisting}
\caption{Example of unsolvable case.}
\label{tab: unsolvable}
\end{figure}
\newpage
\section{Examples and Discussions}
\label{app:tom_examples}
We show examples of good and bad cases of metal states generated by ToMPro and make the following observations:

\begin{itemize}[itemsep=0pt,topsep=0pt]
    \item In general, we found that the GPT-4 is good at understanding the emotions of the fictional characters. Figure \ref{fig:ex1_input} and \ref{fig:ex1_good_output} show an example of an input scene and its output states, where the emotion description is not only accurate but also comprehensive.
    \item In most of the cases, the generated desires are bad, due to the lack of history information as discussed at the end of Section~\ref{ssec:analysis}. The correct desires of characters are usually not reflected in one scene but require reasoning through a sequence of important events. Figure~\ref{fig:ex1_bad_output} gives an example of the bad cases for the same input scene of Figure \ref{fig:ex1_input}.
    \item Another type of bad case is where the GPT-4 tends to follow shallow text cues and output non-informative and misleading descriptions. Figure~\ref{fig:ex1_bad_output} shows an example along the \emph{belief} dimension. Here believing \emph{himself not an alien} is not part of Hulk's thoughts. Though the fact is not incorrect, including such information in the mental states deviates the persona of Hulk, which further misleads the guessing model when predicting the identities. A similar problem exists in the part of \emph{regaining some dignity} of the  \emph{intention} dimension. This highlights a fundamental challenge faced by LLMs trained using co-occurrence-based objectives: distinguishing what is important from the unimportant ones still remains difficult.
    \item Even with access to the history, the GPT-4 may still make mistakes in understanding the personalities, as shown in the example in Figure \ref{fig:ex_err_68i} and \ref{fig:ex_err_68o}, where the character Taylor (Tank) is mistakenly portrayed from a crew member and technical expert to a leader. This will lead to error propagation in the iterative generation process, which explains why incorporating the iterative generation into the other four dimensions hurt the performance, due to the existence of misleading information shown in the previous bullet.
    \item Finally and most crucially, even for most of the scenes, GPT-4 can generate mental descriptions that are majorly correct and meaningful, but the guessing model still cannot make correct predictions. There are two reasons: (1) Insufficient ToM reasoning capabilities: The generated mental states along the dimensions like belief, intention, and desire are usually described with specific details that may not always align with the events in a testing scene. While humans can establish connections between disparate events and induce abstract patterns of thoughts, GPT-4 struggles in this regard.
    (2) Insufficient global understanding: A commonly observed issue is the lack of contextual coherence between the current testing scene and the immediate thoughts stemming from preceding scenes (Figure \ref{fig:ex_global_80}). In such scenarios, it requires either locating relevant mental states that originated several scenes ago or synthesizing the character's cognitive processes from a multitude of past states. This global perspective remains absent in LLMs.
\end{itemize}

\newpage
\begin{figure}[H]
    \centering
\begin{lstlisting}[language=ioexample]
<<<[Scene]>>> in ---The Avengers--- ([[[ALEX]]] -- Hulk)
:::Background:::: ---[INT. ABANDONED BUILDING - DAY]---
:::Background:::: ---[Alex wakes up, in human form. He looks around and sees he is IN A PILE OF RUBBLE and looks up at the OPEN CEILING HE CRASHED IN. AND IS COMPLETELY NAKED. A SECURITY GUARD stands there, amazed.
SECURITY GUARD: You fell out of the sky.]---
[[[ALEX]]]: Did I hurt anybody? 
:::SECURITY GUARD:::: There's nobody around here to get hurt. You did scare the hell out of some pigeons though. 
[[[ALEX]]]: Lucky. 
:::SECURITY GUARD:::: Or just good aim. You were awake when you fell. 
[[[ALEX]]]: You saw? 
:::SECURITY GUARD:::: The whole thing, right through the ceiling. Big and green and buck ass nude. Here... 
:::Background:::: ---[He throws Alex a pair of big pants. Alex pulls on his pants.]---
:::SECURITY GUARD:::: I didn't think those would fit you until you shrunk down to a regular size fella. 
[[[ALEX]]]: Thank you. 
:::Background:::: ---[A beat.]---
:::SECURITY GUARD:::: Are you an alien? 
[[[ALEX]]]: What? 
:::SECURITY GUARD:::: From outer space, an alien? 
[[[ALEX]]]: No. 
:::SECURITY GUARD:::: Well then, son, you've got a condition. 
\end{lstlisting}
    \vspace{-\baselineskip}
    \caption{Example of the input scenes for ToMPro in \textit{The Avengers}.}
    \label{fig:ex1_input}
\end{figure}
\begin{figure}[H]
    \centering
\begin{lstlisting}[language=ioexample]
<<<[Mental States]>>> of [[[ALEX]]] -- Hulk
|||[personality]||| Alex's equilibrium of tranquil demeanor and profound medical knowledge persists, now coupled with an unexpected resilience to physical transformation and the capacity to maintain ethical integrity in the most bewildering circumstances. His ability to remain composed and concerned for the safety of others, even when faced with personal disarray, underscores his dedication to community welfare. However, this scene reveals a heightened vulnerability to extraordinary stress, which can manifest in a loss of physical control, adding a layer of complexity to his character. While his adaptability to drastic changes is commendable, it also introduces an element of unpredictability that could compromise his medical professionalism. Alex must navigate this newfound volatility while safeguarding his moral compass, as his emotional involvement remains a double-edged sword that can both inspire and cloud his judgment. His tendency toward impulsiveness in the face of ethical dilemmas is a weakness that continues to challenge the balance between his personal identity and his professional obligations.
|||[emotions]||| Alex likely feels a mixture of confusion and relief upon waking up. His confusion stems from the unfamiliar and potentially dangerous situation of being in a pile of rubble and naked. Relief may come from the security guard's assurance that no one was hurt. The brief expression "Lucky" indicates a possible fleeting feeling of gratitude or relief. When questioned about being an alien, his response, "What?" may indicate surprise or disbelief.
\end{lstlisting}
    \vspace{-\baselineskip}
    \caption{Example of the good cases of the mental states generated by ToMPro in \textit{The Avengers}.}
    \label{fig:ex1_good_output}
\end{figure}
\begin{figure}[H]
    \centering
\begin{lstlisting}[language=ioexample]
<<<[Mental States]>>> of [[[ALEX]]] -- Hulk
|||[beliefs]||| Alex seems to believe that he could have caused harm, as his first question upon gaining consciousness is about the safety of others. He does not believe he is an alien, as indicated by his response to the security guard's question.
|||[intentions]||| Alex's intention upon waking seems to be to understand what happened to him and to ensure that his actions did not harm anyone. His acceptance of the pants from the security guard indicates an intention to address his immediate need for clothing and to regain some dignity.
|||[desires]||| Alex desires to remain unharmed and to ensure others are safe, as evidenced by his concern about potentially hurting someone. He also desires to cover himself, shown by his quick action to put on the pants provided by the security guard. Alex likely desires to understand his situation better and to resolve any issues arising from his condition, as suggested by the security guard's final comment.
\end{lstlisting}
    \vspace{-\baselineskip}
    \caption{Example of the bad cases of mental states generated by ToMPro from the same scene in \textit{The Avengers}.}
    \label{fig:ex1_bad_output}
\end{figure}

\newpage
\begin{figure}[H]
    \centering
\begin{lstlisting}[language=ioexample]
|||[Character Descriptions]|||
[[[MARY]]]: Mary's <<<leadership>>> is exemplified by her unwavering commitment to her team's safety and success, even at great personal risk, as she demonstrates the ability to prioritize critical objectives under extreme pressure. Her calm demeanor in the face of danger reinforces her role as a stabilizing force, while her nurturing philosophy and clear direction remain central to her approach. Mary's interrogation skills are essential in navigating complex situations. However, her preference for privacy and control can lead to inflexibility when unpredictability arises, a trait that may compromise her leadership in moments requiring swift action. Her empathetic nature fosters trust among her team, yet her cautiousness in high-stakes scenarios is a vulnerability that could hinder decisive decision-making in urgent circumstances.
[[[ALEX]]]: Alex's <<<leadership>>> continues to be defined by his willingness to make sacrifices and his readiness to take decisive action, especially in moments of crisis. His innovative problem-solving abilities, such as utilizing unconventional items for defense, highlight his resourcefulness and strategic acumen. The scene underscores Alex's resilience and mental fortitude, as he faces overwhelming odds with a smile and an unyielding spirit, indicating an inner strength that complements his physical combat skills. His adaptability and quick reflexes, essential in combat, are matched by his composure under pressure, allowing him to remain focused and tactically astute even in dire situations. Nonetheless, Alex's emotional detachment is a persistent shortcoming, creating a barrier to deeper connections with his team and potentially affecting group unity and morale. Despite his effective communication and skillful resource management, the challenge for Alex is to overcome his emotional reserve to achieve a more cohesive team dynamic. His weaknesses include a tendency towards emotional detachment and a possible overreliance on personal strength in situations that may call for collective effort.
[[[JORDAN]]]: Jordan's intellectual rigor and analytical nature are now punctuated by his heightened perception, as he demonstrates an ability to detect and ponder the significance of irregularities, such as the phenomenon of deja vu. His blend of logic and intuition is refined by a cautious openness to the mystical, which he navigates without abandoning his inherent skepticism. This skepticism, however, may occasionally hinder his acceptance of inexplicable events, revealing a tension between his desire for empirical understanding and the enigmatic realities he encounters. Jordan's journey is marked by a struggle to reconcile his quest for knowledge with the acceptance of the unfathomable, with his reluctance to fully embrace such mysteries presenting a consistent challenge to his personal and intellectual growth.
[[[TAYLOR]]]: Taylor's exuberance and cultural flair continue to be integral to his approach, as he displays a keen operational awareness and an ability to swiftly diagnose and respond to emergent challenges with confidence. His quick identification of nerve gas in a critical situation underscores his practical knowledge and readiness to act, reinforcing his potential as a leader. However, his impulsiveness remains a double-edged sword; while it enables rapid response, it also risks precipitous decisions that may not always align with a well-considered strategy. His mentorship is spirited and effective, yet his challenge lies in tempering his instinctive reactions with more deliberate planning to avoid jeopardizing the team's long-term objectives. Taylor's weakness is his propensity to prioritize immediate action over strategic depth, which could lead to oversight in complex operations.

<<<[Scene]>>> in ---The Matrix--- ([[[MARY]]] -- Trinity, [[[ALEX]]] -- Morpheus, [[[JORDAN]]] -- Neo, [[[TAYLOR]]] -- Tank)
:::Background:::: ---[INT. MAIN DECK]---
:::Background:::: ---[Taylor reaches out to the screen as if reaching for Alex.]---
:::TAYLOR:::: No! 
\end{lstlisting}
    \vspace{-\baselineskip}
    \caption{Example of the input scene for ToMPro in \textit{The Matrix}.}
    \label{fig:ex_err_68i}
\end{figure}
\begin{figure}[H]
    \centering
\begin{lstlisting}[language=ioexample]
|||[personality]||| of [[[TAYLOR]]] -- Tank
Taylor's innate <<<leadership>>> potential shines through his spirited nature and cultural vibrancy, complemented by a sharp operational acuity that allows him to quickly recognize and address immediate threats. His proficiency in detecting and reacting to critical situations, such as identifying nerve gas, is a testament to his practical expertise and readiness for action. However, his impulsive tendencies continue to be a liability; his instinct-driven decisions, while swift, may bypass the necessary strategic considerations for long-term success. Taylor's mentorship is enthusiastic and impactful, but he must learn to balance his quick reflexes with thoughtful deliberation to enhance his leadership capabilities and ensure the safety and effectiveness of his team. His weakness lies in his inclination to act first and plan later, a trait that could compromise the intricacy required in multi-faceted missions.
// This error is propagated to all subsequent scenes.
\end{lstlisting}
    \vspace{-\baselineskip}
    \caption{Example of the bad cases of personality generated by ToMPro from an information-poor scene in \textit{The Matrix}.}
    \label{fig:ex_err_68o}
\end{figure}

\newpage
\begin{figure}[H]
    \centering
\begin{lstlisting}[language=ioexample]
<<<[Scene]>>> in ---The Matrix--- ([[[P0-Jordan]]] -- Neo, [[[P1-Taylor]]] -- Tank, [[[P2-Mary]]] -- Trinity)
:::Background:::: ---[INT. MAIN DECK]---
:::Background:::: ---[P0 looks at Alex whose body is covered with a cold sweat.]---
[[[P0]]]: What are they doing to him? 
[[[P1]]]: They're cracking his mind. 
[[[P0]]]: How? 
[[[P1]]]: They inject virus-like serums to break down the system.  It's like cracking a computer.  All it takes is time. 
[[[P0]]]: How much time? 
[[[P1]]]: Depends on the mind.  But eventually, it will crack and his alpha pattern will change from this to this. 
:::Background:::: ---[P1 punches several commands on Alex' personal unit. The monitor waves change from a chaotic pattern to an orderly symmetrical one.]---
[[[P1]]]: When it does, Alex will tell them anything they want to know. 
[[[P0]]]: The access codes to Zion. 
[[[P1]]]: If an agent got inside Zion's mainframe he could do anything. Disable the defense system. It would be the end of us. 
:::Background:::: ---[He looks up at P2 who is pacing relentlessly.]---
[[[P1]]]: We can't let that happen.  We have to do it, P2.  Zion has to be protected. 
:::Background:::: ---[P2 sees Cypher's dead body. Rage overtakes her and she starts kicking hin.]---
[[[P2]]]: Goddamnit!  Goddamnit! 
[[[P1]]]: We have to pull the plug. 
[[[P2]]]: No! 
[[[P1]]]: We don't have any other choice. 
:::Background:::: ---[Those words are like using gasoline to put out a fire and we watch the pain in her eyes burn into a blaze. She walks past him and gets into her chair.]---
[[[P1]]]: P2, what are you doing? 
[[[P2]]]: I'm going in after him. 
[[[P1]]]: Alex could conform at any minute -- 
[[[P2]]]: If he does I'm sure you'll do what has to be done. 
[[[P1]]]: You saw that place.  It's suicide. 
:::Background:::: ---[She glares at him.]---
[[[P1]]]: I know what Alex means to you -- 
[[[P2]]]: No.  No, I don't think you do.  If you did, then you would know that I'm not letting him go.  Not without a fight, P1. 
:::Background:::: ---[She yanks her harness tight.]---
[[[P2]]]: Not without a fight.
// Model Prediction: P0-Mary, P1-Jordan, P2-Taylor
<<<[Options]>>>
(A) [[[Taylor]]] |||[personality]||| Taylor's leadership shines through his ability to take swift and decisive action, with his strategic mindset and situational awareness guiding him in commanding his team effectively. His quick thinking and operational leadership are enhanced by a vibrant personality and cultural savvy. However, his scene reveals a vulnerability to physical harm, highlighting his mortal limitations despite his strong leadership facade. Taylor's enthusiasm for mentorship and imparting practical knowledge is a strength, yet his tendency toward impetuousness can lead to hasty decisions that may bypass deeper strategic considerations. His challenge is to temper his instinctual responses with a more measured, analytical approach to ensure comprehensive mission success. Taylor's weakness is his impulsiveness, which could potentially compromise the well-thought-out strategies necessary in complex operations. |||[emotions]||| Taylor likely feels pain from his injuries, indicated by his charred wounds, but he may also be experiencing emotional distress, possibly due to something that happened to Dozer. The tightening of his face upon Mary's inquiry about Dozer suggests he might be feeling sadness, guilt, or anger. |||[beliefs]||| Taylor seems to believe that he can manage his injuries, as indicated by his reassurance to Mary that he'll be alright. His belief may also extend to a sense of duty or responsibility, which is why he might be downplaying his own pain. |||[intentions]||| Taylor's immediate intention appears to be to reassure Mary about his condition, possibly to keep her from worrying or to maintain a sense of normalcy in a stressful situation. His response to the mention of Dozer suggests he might be intending to avoid the topic, possibly because it's painful or he needs to focus on the present situation. |||[desires]||| Taylor desires to appear strong and capable, perhaps to maintain his own morale or that of Mary. He may also have a desire to protect Mary from further distress, which could be why he doesn't elaborate on his condition or the situation with Dozer. His desire could also include wishing to address his injuries and deal with the aftermath of whatever happened to Dozer, but these are secondary to his immediate interaction with Mary.
(B) [[[Alex]]] {{ omitted }} (C) [[[Mary]]] {{ omitted }} (D) [[[Jordan]]] {{ omitted }}
// The generated ToMs contain no information that Taylor is a technical expert.
\end{lstlisting}
    \vspace{-\baselineskip}
    \caption{Example of the bad cases of insufficient global understanding in \textit{The Matrix}.}
    \label{fig:ex_global_80}
\end{figure}
\newpage

\section{Additional Experimental Results}
\label{app:addditional_exp}

\paragraph{Performance Breakdown to Number of Speakers in a Scene}
Table~\ref{tab:number_choices_results} presents the performance breakdown of different approaches. We make similar observations in Table~\ref{tab:breakdown_difficulty}, where humans and our ToMPro approach lead to smaller performance differences across different numbers of speakers.

\begin{table}[h!]
\centering
\caption{Performance decomposition to the number of choices in a scene. (*) Conducted on the subset of human evaluation. There is only one 5-speaker scene, thus the number of 100\% is not significant.}
\vspace{-0.7\baselineskip}
\begin{tabular}{ccccccc} 
\toprule
     \multirow{2}{*}{\textbf{\#Speakers}} &
     \multicolumn{2}{c}{\bf Transductive} &
     \multicolumn{2}{c}{\bf Inductive} &
     \multirow{2}{*}{\bf Human$^*$} \\
     \cmidrule(lr){2-3} \cmidrule(lr){4-5}
     & \textbf{ProtoNet}
     & \textbf{GPT-4 ICL$^*$}
     & \textbf{LEOPARD}
     & \textbf{GPT-4 ToMPro$^*$} \\
\midrule
 1 & 56.9 & 81.1 & 65.5 & 74.2 & 93.4 \\
 2 & 55.3 & 64.9 & 61.3 & 67.4 & 86.5 \\
 3 & 50.2 & 62.5 & 53.7 & 62.5 & 84.8 \\
 4 & 43.0 & 55.4 & 41.5 & 71.9 & 82.1 \\
 5 & 46.1 & 22.0 & 36.5 & 16.7 & 90.0 \\
 \bottomrule
\end{tabular}
\label{tab:number_choices_results}
\end{table}

\paragraph{Performance Breakdown to Movie Genres}
Table~\ref{tab:breakdown_genre} details the performance breakdown by movie genre across different methods.

\begin{table}[h!]
\centering
\caption{Performance decomposition to movie genres. (*) Conducted on the subset of human evaluation.}
\vspace{-0.7\baselineskip}
\begin{tabular}{llccccc} 
\toprule
     &
     \multirow{2}{*}{\textbf{Genre}} &
     \multicolumn{2}{c}{\bf Transductive} &
     \multicolumn{2}{c}{\bf Inductive} &
     \multirow{2}{*}{\bf Human$^*$} \\
     \cmidrule(lr){3-4} \cmidrule(lr){5-6}
     &
     & \textbf{ProtoNet}
     & \textbf{GPT-4 ICL$^*$}
     & \textbf{LEOPARD}
     & \textbf{GPT-4 ToMPro$^*$} \\
\midrule
 & Action    & 53.8 & 71.1 & 59.7 & 71.1 & 86.0 \\
 & Adventure & 61.3 & 78.7 & 68.0 & 90.0 & 98.4 \\
 & Comedy    & 48.5 & 45.7 & 51.8 & 51.0 & 86.3 \\
 & Crime     & 69.6 & 85.0 & 80.4 & 82.1 & 93.3 \\
 & Drama     & 58.5 & 49.2 & 71.7 & 60.7 & 91.8 \\
 & Horror    & 66.7 & 64.7 & 64.1 & 68.5 & 77.6 \\
 & Romance   & 52.4 & 92.0 & 61.0 & 94.7 & 84.0 \\
 & Sci-Fi    & 50.7 & 60.3 & 45.1 & 47.8 & 88.3 \\
 & Thriller  & 55.1 & 70.0 & 59.3 & 53.9 & 90.0 \\
\bottomrule
\end{tabular}
\label{tab:breakdown_genre}
\end{table}

\paragraph{Subset of Testing Movies used in Table~\ref{tab:overall_performance}}
To keep the samples covering similar genres like in our development subset~Table~\ref{tab:dev_sampled_movies}, we sample the testing movies following Table~\ref{tab:test_sampled_movies}.

\begin{table}[H]
\caption{Sampled test set by genre.}
\vspace{-0.7\baselineskip}
\centering
\begin{tabular}{llcc} 
\toprule                
\bf Movie Genre &
\bf Example     &
\bf \#Scene       &
\bf \#Instances    \\
\midrule
 Action    & \emph{Terminator Salvation}          & 26 & 40 \\
 Adventure & \emph{The Avengers}                  & 14 & 17 \\
 Comedy    & \emph{American Pie}                  & 30 & 46 \\
 Crime     & \emph{Catch Me If You Can}           & 41 & 45 \\
 Drama     & \emph{Precious}                      & 30 & 33 \\
 Horror    & \emph{A Nightmare On Elm Street}     & 19 & 26 \\
 Romance   & \emph{Passengers}                    &  7 & 15 \\
 Sci-Fi    & \emph{The Matrix}                    & 39 & 67 \\
 Thriller  & \emph{Donnie Brasco}                 & 30 & 47 \\
\bottomrule
\end{tabular}
\label{tab:test_sampled_movies}
\end{table}

\section{Details of Character Memory Solution with LLMs}
\label{app:mem_pro}
A straightforward inductive learning approach is to retrieve LLMs' memory of characters to text descriptions (with the prompt in Figure~\ref{app:gpt-memory}), then feed the descriptions as the character representation to LLMs for identity guessing (with the prompt in Figure~\ref{fig:llm-methods}b).

The performance of this method could be considered as the cap for the inductive approaches, as the generated text descriptions about characters contain spoiler information, which is in favor of fulfilling the evaluation tasks performed on movie endings.

\begin{figure}[H]
    \centering
    \begin{lstlisting}[language=prompt]
Please summarize the personality and traits of the character [[[CHAR]]] in the movie [[[MOVIE_NAME]]] in a single paragraph.
    \end{lstlisting}
    \caption{Prompt template for retrieving LLMs' memory.}
    \label{app:gpt-memory}
\end{figure}

\end{document}